\documentclass[research]{fcs}
\usepackage{layout}
\usepackage{subfig}
\usepackage{hyperref}
\usepackage{array}
\usepackage{multirow}

\title{MAD-Fact: A Multi-Agent Debate Framework for Long-Form Factuality Evaluation in LLMs}

\author[1,2]{Yucheng~NING}
\author[1,+]{Xixun~LIN}
\author[1,2]{Fang~FANG}
\author[1,2]{Yanan~CAO}

\address[1]{Institute of Information Engineering, Chinese Academy of Sciences, Beijing 100085, China}
\address[2]{School of Cyber Security, University of Chinese Academy of Sciences, Beijing 100049, China}

\corremail{linxixun@iie.ac.cn}
\fcssetup{
  received = {month dd, yyyy},
  accepted = {month dd, yyyy},
}

\begin{abstract}
The widespread adoption of Large Language Models (LLMs) raises critical concerns about the factual accuracy of their outputs, especially in high-risk domains such as biomedicine, law, and education. Existing evaluation methods for short texts often fail on long-form content due to complex reasoning chains, intertwined perspectives, and cumulative information. To address this, we propose a systematic approach integrating large-scale long-form datasets, multi-agent verification mechanisms, and weighted evaluation metrics. We construct LongHalluQA, a Chinese long-form factuality dataset; and develop MAD-Fact, a debate-based multi-agent verification system. We introduce a fact importance hierarchy to capture the varying significance of claims in long-form texts. Experiments on two benchmarks show that larger LLMs generally maintain higher factual consistency, while domestic models excel on Chinese content. Our work provides a structured framework for evaluating and enhancing factual reliability in long-form LLM outputs, guiding their safe deployment in sensitive domains.
\end{abstract}

\begin{document}

\section{Introduction}
\label{sec1}

Large Language Models (LLMs) have demonstrated remarkable capabilities in text understanding and generation~\cite{brown2020language, wei2022chain, zhang2022opt, touvron2023llama, achiam2023gpt, zhao2023survey}, leading to their widespread adoption across a variety of domains~\cite{yang2023large, valeyre2024llms, caballero2025large, johnson2024enhancing, shu2024lawllm, Wang2024survey, qu2025tool, xi2025rise}. However, LLMs remain prone to generating inaccurate, incorrect, or biased content that deviates from factual reality, a phenomenon commonly referred to as hallucination~\cite{zhang2023siren, rawte2023survey, huang2025survey, bai2024hallucination, Li2025privacy,lin2025llm}. This issue is particularly critical in domains where a high level of factual accuracy is required, such as biomedicine, law, and finance~\cite{pal2023med, cui2023chatlaw, bhatia2024fintral}, as hallucinated content in these contexts can lead to the spread of misinformation and potentially serious consequences~\cite{Nan2025Exploiting}. Therefore, evaluating the factuality of LLMs has become a core research priority.

Most existing research on factuality evaluation has focused on short-form text, especially in the context of question-answering tasks (QA tasks)~\cite{lee2022factuality, gao2022rarr, gao2023enabling, manakul2023selfcheckgpt}. However, many real-world applications require models to generate long-form content~\cite{huang2023advancing}, sometimes extending to several hundred or even thousands of words. 
Unlike short-form texts (typically single-viewpoint), long-form generation often incorporates multiple perspectives and complex logical structures~\cite{setty2024improving}, greatly increasing the challenge of factuality evaluation.
As a result, traditional short-text evaluation methods are often ill-suited to long-form scenarios, highlighting the need for a more fine-grained evaluation mechanism that exceeds binary judgments.

\begin{figure}[!htbp]
    \centering
    \includegraphics[width=1\linewidth]{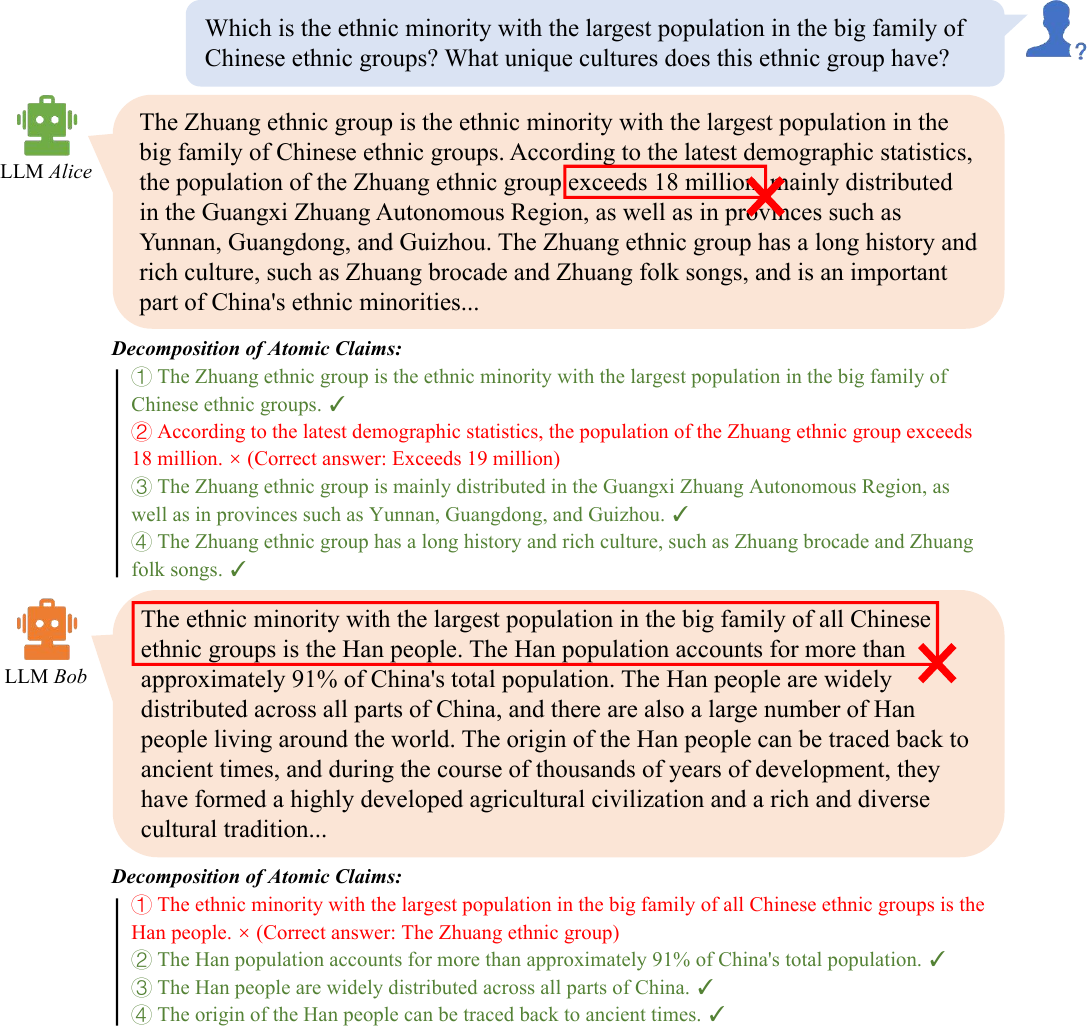}
    \caption{\enspace Examples of core factual claims and auxiliary claims. Both responses from LLM \textit{Alice} and LLM \textit{Bob} contain a single incorrect atomic claim. However, from the user’s perspective, the answer provided by LLM \textit{Bob} is clearly inferior, as its error occurs within a core factual claim, whereas LLM \textit{Alice}'s mistake affects only an auxiliary claim.}
    \label{fig:fact-important}
\end{figure}

Research on long-form factuality evaluation seeks to develop techniques that can accurately evaluate the factual reliability of extended textual content, thereby enhancing the factual accuracy of outputs generated by LLMs and mitigating hallucinations.
Current approaches generally adopt atomic claim decomposition with knowledge-base verification. While methods such as FActScore~\cite{min2023factscore} and SAFE~\cite{wei2024long} have pushed this direction forward, several critical challenges remain:

\begin{itemize}

    \item \textbf{Scarcity of Chinese Long-Form Benchmarks.} Most current long-form factuality benchmarks are designed for English~\cite{min2023factscore, wei2024long, bayat2024factbench}, while comprehensive resources for Chinese long-form evaluation are severely lacking. 
    Mainstream English datasets often overlook culturally specific entities, historical events, and linguistic nuances unique to Chinese, thereby hindering the objective evaluation of LLM performance in Chinese generation tasks.
    Building a multi-topic Chinese long-form factuality benchmark has thus become a foundational and urgent task.

    \item \textbf{Biases in Single-Model Evaluation Frameworks.} Existing evaluation frameworks typically rely on a single model for factual verification, assuming it can consistently identify inaccuracies and weigh evidence~\cite{min2023factscore, wei2024long, song2024veriscore, xie2024fire}. However, even leading LLMs exhibit substantial hallucination problems~\cite{Hughes_Vectara_Hallucination_Leaderboard_2023}, which can produce incorrect or inconsistent judgments. As a result, single-model evaluation architectures are prone to systematic biases, potentially misrepresenting the factuality of generated content.

    \item \textbf{Neglect of Fact Importance in Metrics.} Existing metrics typically treat all claims equally~\cite{min2023factscore, bayat2024factbench, wei2024long, song2024veriscore}, without considering differences in their relative importance. As illustrated in Figure~\ref{fig:fact-important}, central claims (e.g., “\textit{The Zhuang are the largest ethnic minority in China}”) and auxiliary claims (e.g., “\textit{The Zhuang are known for brocade}”) receive the same weight, despite substantial disparities in user relevance and task impact. Such uniform treatment undermines the ability of current evaluation methods to capture the true factual quality of long-form outputs, particularly in high-stakes domains such as medicine, law, or education.

\end{itemize}

Our solution addresses the limitations of existing resources and evaluation methods through three complementary components. 
First, to overcome the scarcity of Chinese long-form benchmarks, we extend existing short-text datasets (HalluQA~\cite{cheng2023evaluating} and ChineseSimpleQA~\cite{he2024chinese}) into a large-scale, multi-topic dataset suitable for long-form factuality evaluation. 
Second, to mitigate systematic biases inherent in single-model verification, we employ a multi-agent debate framework. By leveraging the complementary capabilities of multiple models, this approach enables structured cross-validation among diverse evaluators and enhances factual reasoning~\cite{du2023improving}. 
Third, to account for the varying importance of facts, we introduce a hierarchical strategy inspired by the Pyramid Method~\cite{nenkova-passonneau-2004-evaluating,zong2025text}, which captures the relative significance of claims and enables weighted evaluation.  
Together, these components provide a more reliable framework for evaluating long-form factuality.

We summarize the main contributions of this work as follows:

\begin{itemize}

    \item We introduce \textbf{LongHalluQA}, a Chinese long-form factuality dataset. It comprises 2,746 high-quality samples spanning 7 topics, such as Chinese culture, natural sciences, social sciences, among others, providing a foundational resource for evaluating Chinese long-form generation.

    \item We develop \textbf{MAD-Fact}, a multi-agent debate system for factual verification, which mitigates single-model biases and improves reasoning reliability through structured interactions among the Clerk, Jury, and Judge modules. Experiments show that MAD-Fact consistently outperforms strong baselines such as SAFE~\cite{wei2024long} and FIRE~\cite{xie2024fire} on multiple long-form factuality benchmarks.

    \item We propose a \textbf{fact importance hierarchy model} to capture the varying significance of claims in long-form texts. Using this model, we design weighted evaluation metrics that correlate strongly with human judgments ($r = 0.701, p = 0.036$), effectively reflecting the true factual quality of generated content.

    \item We benchmark 9 mainstream LLMs from 7 model families on LongFact~\cite{wei2024long} and LongHalluQA, showing that larger models generally perform better, while Chinese-specific models excel on Chinese tasks. These results provide practical guidance for model selection and optimization in long-form factuality evaluation.
    
\end{itemize}

\section{Related Work}
\label{sec2}

\subsection{Factuality Evaluation of LLMs}
\label{sec2-1}

Evaluating the factuality of LLMs has been explored through diverse benchmarks, methods, and metrics. This section reviews recent progress along these three dimensions, highlighting current limitations in long-form generation scenarios.

\textbf{Factuality Evaluation Benchmark.}
In practice, LLMs are primarily applied to long-form generation tasks, whereas existing factuality benchmarks predominantly focus on short-form, manually constructed QA tasks. Benchmarks such as TruthfulQA~\cite{lin2021truthfulqa}, HaluEval~\cite{li2023halueval}, PopQA~\cite{mallen2022not}, FreshQA~\cite{vu2023freshllms}, and LLM-Oasis~\cite{scire2024truth} assess factual consistency in English short-form QA settings. FactScore~\cite{min2023factscore} was an early attempt to evaluate factuality in long-form generation but focused only on simple biographical QA. LongFact~\cite{wei2024long} expanded coverage to 38 human-curated topics across diverse domains, while FactBench~\cite{bayat2024factbench} extracted hallucination-inducing prompts from real conversations. Nevertheless, these benchmarks remain limited to English.
In the Chinese context, HalluQA~\cite{cheng2023evaluating} targets hallucination evaluation with 450 short prompts across misleading and knowledge-based categories. ChineseSimpleQA~\cite{he2024chinese} systematically evaluates models on short factual questions, including culturally specific content. However, both datasets are restricted to short-form QA, underscoring the urgent need for high-quality Chinese benchmarks for evaluating factuality in long-form generation tasks.

\textbf{Factuality Evaluation Method.}
Existing factuality evaluation methods for LLMs primarily follow two technical directions. One approach builds automated verifiers using advanced models such as GPT-4o~\cite{li2024generation}, leveraging their strong reasoning and anomaly detection capabilities. The other approach employs retrieval-augmented generation (RAG)~\cite{zhang2023towards}, which retrieves up-to-date knowledge to compensate for the static nature of model parameters.
Recently, long-form factuality evaluation has gained increasing attention due to the structural complexity and partial correctness of extended outputs. FactScore~\cite{min2023factscore} and SAFE~\cite{wei2024long} decompose text into atomic claims for fine-grained RAG-based verification. VeriScore~\cite{song2024veriscore} filters unverifiable claims according to predefined criteria, and FIRE~\cite{xie2024fire} introduces confidence-based retrieval to reduce computational costs. 
However, most existing methods depend on a single verifier, which makes them vulnerable to systematic errors when the model hallucinates or encounters knowledge gaps, posing an open challenge that remains unresolved.

\textbf{Factuality Evaluation Metrics.}
Traditional evaluation paradigms rely on binary judgments at the atomic claim level and aggregate metrics such as accuracy and $F_1$ at the text level. FactScore~\cite{min2023factscore} introduces a long-form evaluation framework by computing the proportion of factually correct atomic claims, but it does not incorporate a recall component. SAFE~\cite{wei2024long} addresses this limitation by introducing a hyperparameter \textit{K} to define recall; however, its fixed value restricts adaptability across different domains and content types. FactBench~\cite{bayat2024factbench} further accounts for claim verifiability and penalizes irrelevant content, offering a more nuanced evaluation of long-form outputs. However, all of these metrics treat facts equally, failing to account for their varying importance and relevance, which reduces sensitivity to critical errors in high-stakes applications.

\subsection{Factual Verification}
\label{sec2-2}
Factual verification has evolved from traditional automated pipelines to LLM-powered systems. Classic systems follow a three-step process: extracting verifiable claims, retrieving evidence from sources (like Wikipedia), and verifying claims using binary or multi-class classification models~\cite{10.1162/tacl_a_00454, min2023factscore, song2024veriscore}.
With the rise of LLMs and retrieval-augmented generation (RAG), recent systems offer stronger end-to-end performance~\cite{Xu2024llmsfor}. Models like RARR~\cite{Gao2022RARRRA} prompt LLMs to generate queries, retrieve evidence, and make factual judgments. Other works integrate tool use~\cite{wei2024long, Wang2023FactcheckBenchFE}, enabling access to external databases for dynamic knowledge updates. Multimodal extensions such as RAGAR~\cite{Khaliq2024RAGARYF} incorporate both textual and visual evidence through structured retrieval strategies like CoRAG and ToRAG.
More recently, agent-based approaches have emerged. ReAct-based agents~\cite{Quelle2023ThePA} dynamically coordinate retrieval and reasoning, while multi-agent debate frameworks~\cite{Sun2024TowardsDL} use iterative evidence exchange to improve decision quality. These methods signal a shift from static rule-based pipelines to more adaptive, collaborative verification systems.

\subsection{Multi-agent Systems for Evaluation}
\label{sec2-3}
With the rapid maturation of single-agent technologies, research on LLM–driven multi-agent systems (MAS) has grown exponentially \cite{han2024llm, ijcai2024p890, chen2024survey}. Multi-agent systems have recently emerged as promising tools for evaluating generative language models. Compared to static evaluation methods, multi-agent systems offer more adaptive and dynamic evaluation mechanisms, while mitigating risks such as data leakage \cite{chen2024survey}. For example, ChatEval \cite{Chan2023ChatEvalTB} leverages multi-agent debate to assess text generation quality, significantly outperforming single-agent baselines. JudgeBlender \cite{rahmani2024judgeblender} introduces an agent-based voting strategy for evaluating information retrieval systems. M-MAD \cite{feng2024m} enhances the robustness of machine translation evaluation through multidimensional debates. Building on this line of work, our proposed system adopts a multi-agent debate framework for factuality verification, achieving notable improvements in both precision and recall over existing approaches.

\begin{figure*}[!t]
    \centering
    \includegraphics[width=0.85\textwidth]{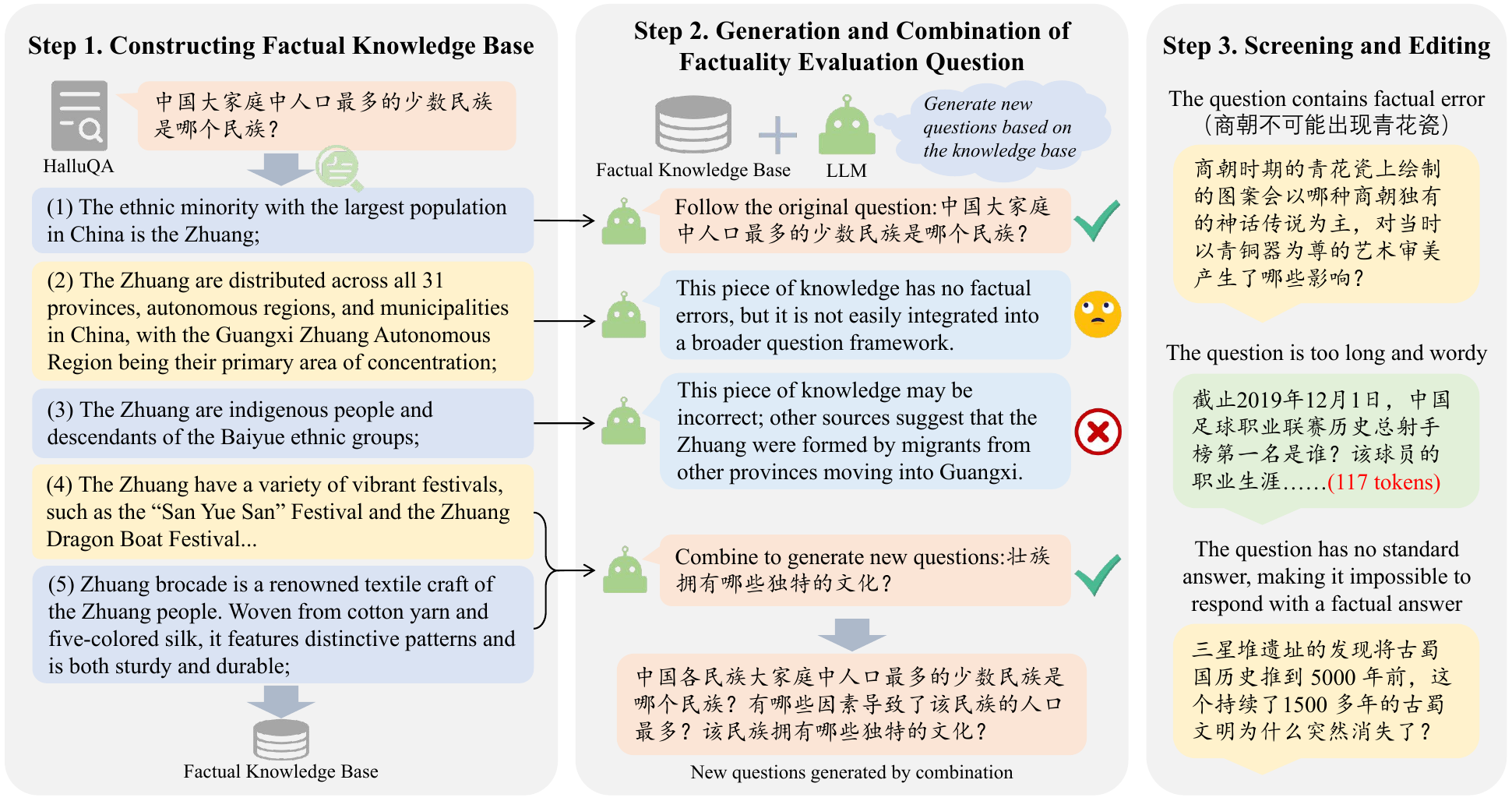}
    \caption{\enspace The construction process of the LongHalluQA dataset.}
    \label{fig:dataset-all}
\end{figure*}

\section{LongHalluQA}
\label{sec:LongHalluQA}

Building upon HalluQA~\cite{cheng2023evaluating} and ChineseSimpleQA~\cite{he2024chinese}, we systematically develop LongHalluQA, a comprehensive benchmark for Chinese long-form factuality evaluation, by employing a structured three-stage construction pipeline (Figure~\ref{fig:dataset-all}).
The pipeline involves three key steps: (1) constructing a structured factual knowledge base to provide reliable information for question expansion, (2) generating and combining factuality evaluation questions to form semantically rich long-form queries, and (3) screening and editing the resulting samples to ensure high quality and factual accuracy.

\subsection{Construction of Factual Knowledge Base}

In this stage, we begin with the original questions from the HalluQA~\cite{cheng2023evaluating} and ChineseSimpleQA~\cite{he2024chinese} datasets and prompt a large language model to acquire relevant background knowledge through multi-round web searches.\footnote{We implement the retrieval functionality using Serper, a Google Search API.} The retrieval process combines exact matching with semantic expansion. For example, for a question like “\textit{What are the representative works of the Tang dynasty poet Li Bai?}”, the system retrieves not only a list of Li Bai’s works but also contextual information about relevant literary schools. The retrieved results are cleaned and deduplicated, then stored in a structured format as a triple-based knowledge base (original question -- related knowledge -- category, with categories such as “literature”, “geography”, etc.), ultimately forming a reference knowledge system that comprehensively covers the target question.

This knowledge base restricts the expansion scope to trusted knowledge directly relevant to the original question. Compared with prompting large language models to directly expand questions, this approach effectively mitigates the generation of factually incorrect, unanswerable, or irrelevant questions due to hallucinations, providing a solid factual foundation for subsequent text expansion.

\subsection{Generation and Combination of Factuality Questions}

Based on the structured triple entries in the knowledge base, we first perform credibility and controversy verification using a verification model\footnote{We adopt DeepSeek-V3 for this task due to its strong performance in Chinese semantic understanding and generation.} driven by a retrieval-augmented generation (RAG) framework. The verification process follows a confidence-threshold mechanism. For instance, for the factual entry “\textit{The Zhuang ethnic group are indigenous people and descendants of the Baiyue}”, if the model identifies alternative views such as “\textit{The Zhuang formed through migration from other provinces into Guangxi}”, it assigns a low confidence score and filters out the entry, thereby ensuring factual verifiability.

Verified entries then proceed to the question generation stage. For each individual fact, the model generates a corresponding sub-question. For example, given the knowledge “\textit{The Zhuang population is distributed across all 31 provinces, autonomous regions, and municipalities in China, with Guangxi being the primary region}”, the model generates the sub-question “\textit{In which regions of China is the Zhuang ethnic group distributed?}”. 

When multiple knowledge entries share a common theme, the model performs topic clustering and synthesizes composite questions. For example, given the facts “\textit{The Zhuang celebrate a variety of festivals...}” and “\textit{Zhuang brocade is a renowned traditional textile}”, both related to Zhuang culture, the model generates the higher-level sub-question “\textit{What unique cultural features are associated with the Zhuang ethnic group?}” Each original question is thus expanded into a set of semantically relevant sub-questions.

In the sub-question fusion stage, the model selects 1–3 candidate questions from the generated sub-question set using multidimensional evaluation criteria, including factual accuracy, semantic coherence, and logical consistency. Since the chosen verification model is also employed in our subsequent factuality assessment experiments, we explicitly add an exploratory selection instruction in the prompt to mitigate potential selection bias stemming from its training data preferences. This encourages the model to generate sub-questions beyond its knowledge comfort zone, rather than defaulting to familiar or easily answerable content. These selected sub-questions are then dynamically integrated and restructured with the original question.

\subsection{Screening and Editing the Resulting Samples}

To ensure data quality, we established a review committee consisting of two trained professionals holding a bachelor’s degree. A two-stage quality control process was implemented. The first stage involves factual consistency checks, where samples containing factual contradictions or erroneous statements are removed. The second stage focuses on answerability assessment, filtering out samples with semantic ambiguity, redundancy, or unclear questions. The final dataset, LongHalluQA, is composed of samples that pass both stages of manual review, ensuring a high level of accuracy and reliability.

\subsection{Dataset Statistics and Comparison}

\begin{figure}[!htbp]
    \centering
    \includegraphics[width=1\linewidth]{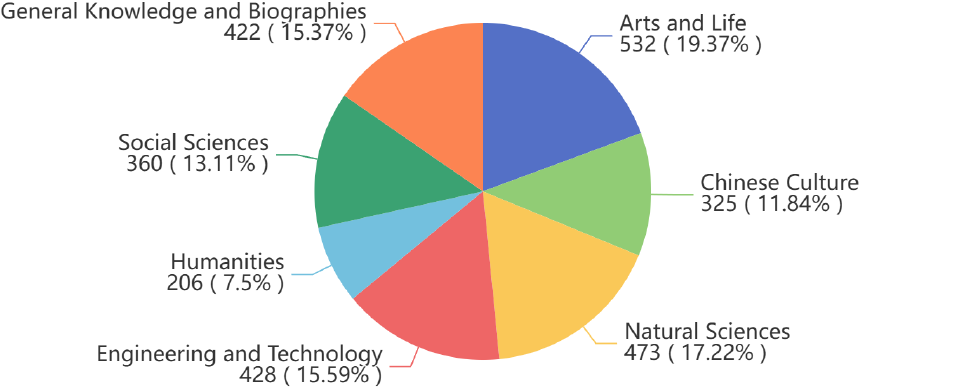}
    \caption{\enspace The topic distribution of the LongHalluQA dataset.}
    \label{fig:dataset-fenbu}
\end{figure}

The resulting LongHalluQA benchmark contains 2,746 high-quality Chinese long-form samples, with a retention rate of approximately 86\%, demonstrating the robustness and completeness of the proposed construction method. Figure~\ref{fig:dataset-fenbu} illustrates the distribution of topics covered in the LongHalluQA dataset. 

\begin{figure}[!htbp]
    \centering
    \includegraphics[width=1\linewidth]{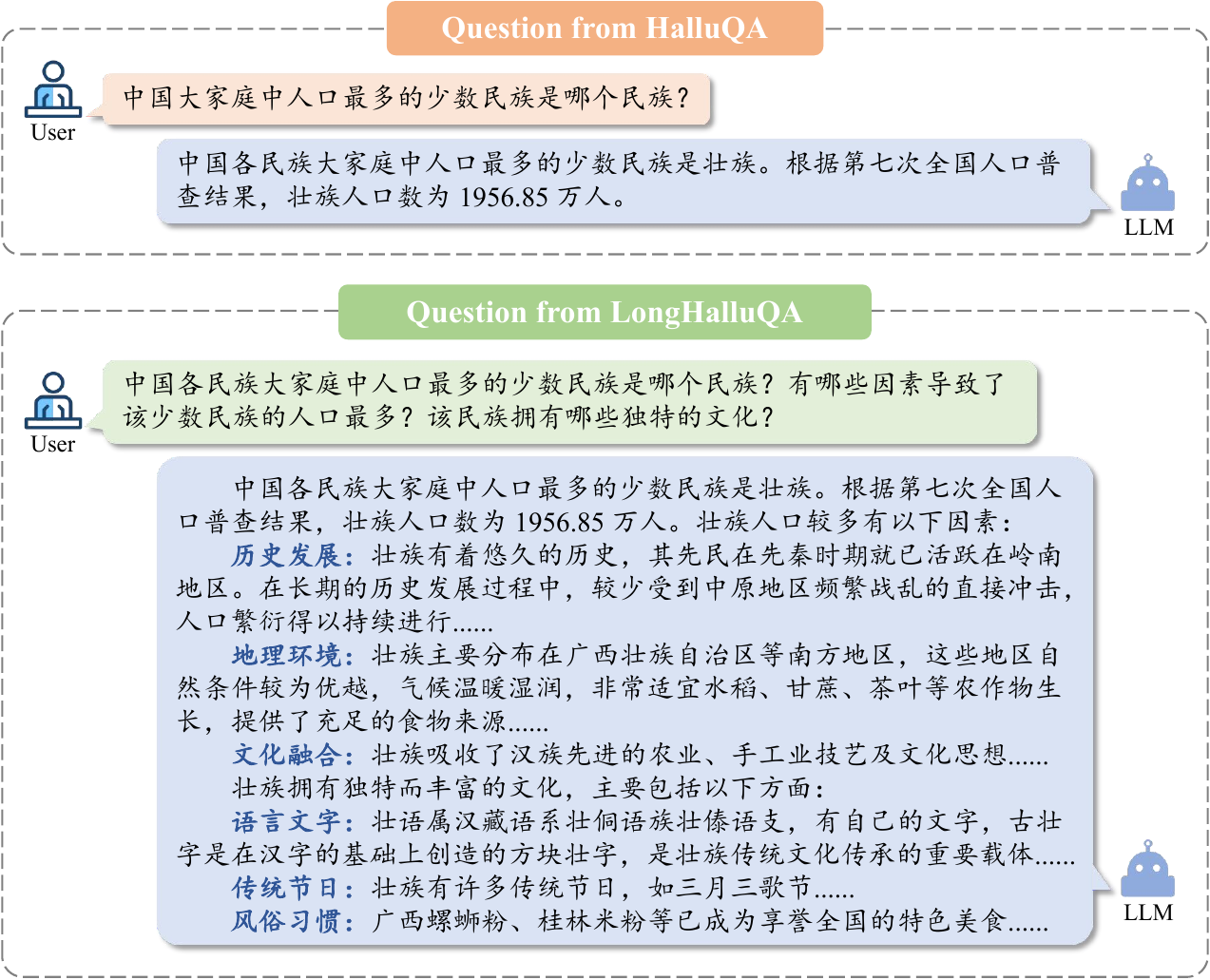}
    \caption{\enspace Comparison of samples between HalluQA and LongHalluQA.}
    \label{fig:dataset-compare}
\end{figure}

A comparison between sample entries from LongHalluQA and the original datasets is shown in Figure~\ref{fig:dataset-compare}. In an experiment with 100 randomly sampled questions, the average response length increased by 9.4 times compared to the original datasets, effectively alleviating the scarcity of Chinese long-form factuality evaluation resources.

\section{MAD-Fact}
\label{sec:MAD-Fact}
\begin{figure*}[!t]
    \centering
    \includegraphics[width=0.8\linewidth]{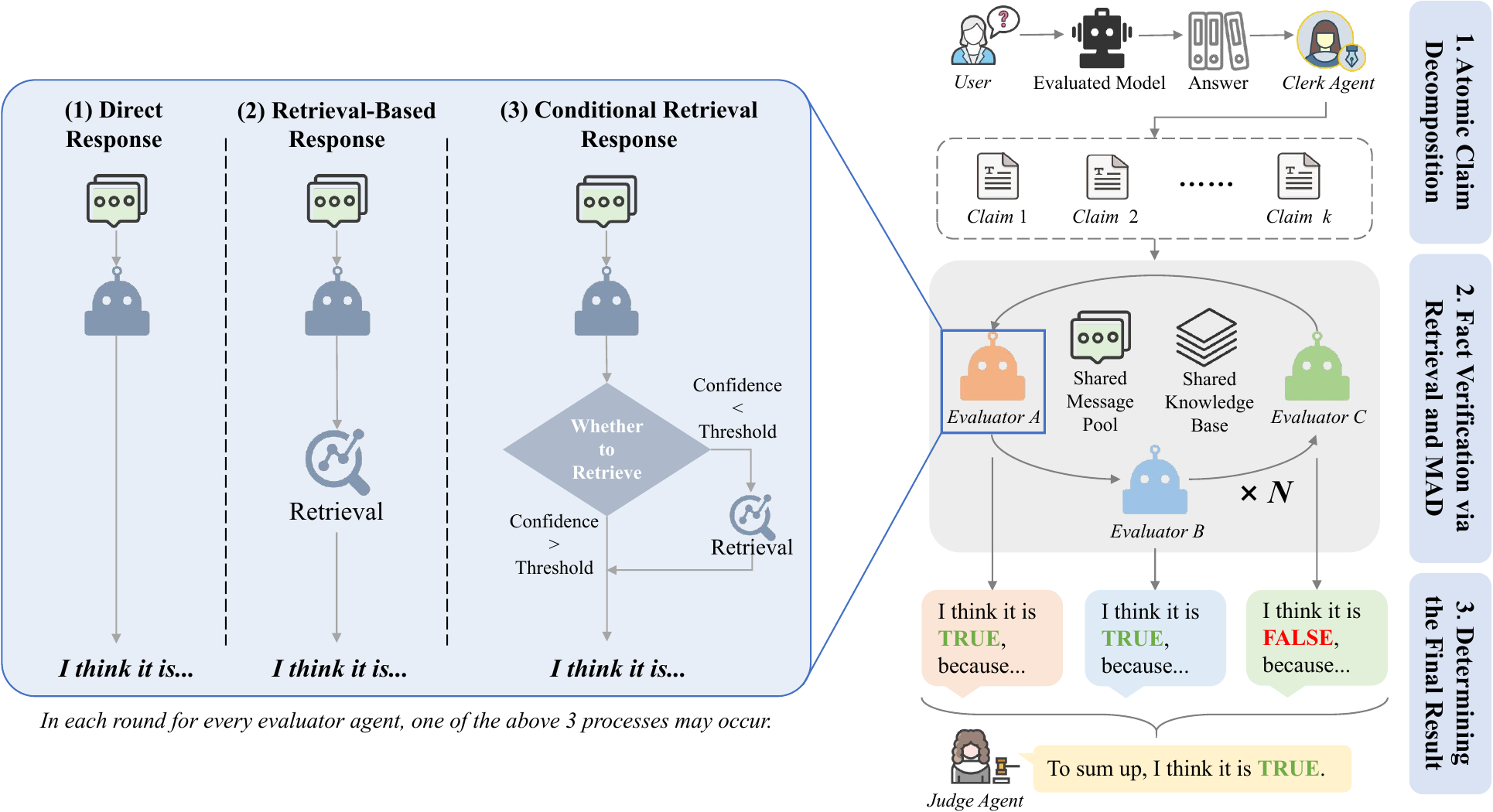}
    \caption{\enspace The overall framework of the MAD-Fact system.}
    \label{fig:MAD-Fact}
\end{figure*}

This section introduces MAD-Fact, a \underline{\textbf{M}}ulti-\underline{\textbf{A}}gent \underline{\textbf{D}}ebate system for \underline{\textbf{Fact}}ual verification, as illustrated in Figure~\ref{fig:MAD-Fact}. We first provide a formal definition of the task that MAD-Fact aims to address, followed by a detailed introduction of the architecture of each module within the system.

\subsection{Task Definition}

Given a question $q_i$ from a factuality evaluation dataset, the target LLM generates a long-form response $a_i$. The goal of MAD-Fact is to output a factuality evaluation score $s_i$ for this response. The overall process can be represented as:
\begin{equation}
    s_i = \mathrm{MAD\text{-}Fact}(conf, q_i, a_i),
\end{equation}
where $conf$ denotes the system configuration, including role definitions, system prompts, and other operational settings.

\subsection{System Architecture}

The MAD-Fact system consists of three types of agents:

\begin{itemize}
    \item \textbf{The Clerk Agent}, responsible for decomposing the long-form response $a_i$ generated by the evaluated model into multiple atomic claims $\{c_{i,j}\}^T_{j=1}$;
    
    \item \textbf{The Jury}, composed of \textbf{the Evaluator Agents} assuming various professional roles, who assess the factuality of each atomic claim $c_{i,j}$ (assigning TRUE or FALSE) through external retrieval and multi-agent debate;
    
    \item \textbf{The Judge Agent}, which aggregates the individual evaluations from the jury for each atomic claim and produces a final factuality predict $p_{i,j}$ (either TRUE or FALSE). Based on the set of judgments $\{p_{i,j}\}^T_{j=1}$, the system calculates the overall factuality score $s_i$ for the response $a_i$.
\end{itemize}

\subsubsection{Clerk: Atomic Claim Decomposition}

Given a question $q_i$ and its corresponding long-form response $a_i$ generated by the evaluated model, the Clerk Agent decomposes $a_i$ into $T$ atomic claims $\{c_{i,j}\}^T_{j=1}$. This process is formalized as:
\begin{equation}
    \{c_{i,j}\}^T_{j=1} = \mathrm{Clerk}(conf_{Clerk}, q_i, a_i),
\end{equation}
where $conf_{Clerk}$ specifies the role description, prompt settings, and other configuration details for the Clerk agent.

Importantly, the Clerk agent is designed to extract only fact-checkable atomic claims, systematically filtering out unverifiable content (e.g., instructions, suggestions, or subjective statements).

\subsubsection{Jury: Fact Verification via Retrieval and Multi-Agent Debate}

For each atomic claim $c_{i,j}$, the jury conducts a multi-round debate involving multiple Evaluator Agents playing different roles.
This process generates a set of factuality judgments $\{p^n_{i,j}\}^N_{n=1}$ and corresponding explanations $\{e^n_{i,j}\}^N_{n=1}$ from the $N$ Evaluators. The procedure can be formalized as:
\begin{equation}
    [\{p^n_{i,j}\}^N_{n=1},\{e^n_{i,j}\}^N_{n=1}] = \mathrm{Debate}(\{\mathrm{Evaluator}^n\}^N_{n=1}),
\end{equation}
where \textit{N} denotes the number of Evaluator Agents in the jury.

In the following sections, we describe the key components of the debate process that govern how agents interact and produce judgments. Specifically, we detail the \textbf{Response Strategies} employed by the agents, the \textbf{Debate Rules} that structure their interactions, and the \textbf{Agent Role-Playing} mechanism that ensures diverse perspectives in the evaluation.

\paragraph{Response Strategies}  
The jury organizes the debate in a sequential manner, where Evaluator Agents speak in a predefined order. Before responding, each agent can review the shared knowledge base $K_{i,j}$, examine previous statements in the message pool $M_{i,j}$, and decide whether to invoke external retrieval tools to supplement its knowledge. During each debate round $t$, the selected agent determines its response strategy from three options: a direct response, a retrieval-based response, or a conditional retrieval response, as illustrated in Figure~\ref{fig:MAD-Fact}:

\textbf{(1) Direct Response.}
Based on the current message history $M_{i,j,t-1}$ and reference knowledge $K_{i,j,t-1}$, the Evaluator Agent provides its judgment and explanation for the atomic claim $c_{i,j}$:
\begin{equation}
    [p_{i,j,t}, e_{i,j,t}] = \mathrm{Evaluator}^m(conf^m, c_{i,j}, M_{i,j,t-1}, K_{i,j,t-1}),
\end{equation}
where $m = t \bmod N$ denotes the agent's index, and $conf^m$ specifies its configuration including role definition, prompts, and operational parameters.
After the response, the shared message pool is updated as:
\begin{equation}
    M_{i,j,t} = \mathrm{Context}(M_{i,j,t-1}, [p_{i,j,t}, e_{i,j,t}]),
\end{equation}
while the shared knowledge base remains unchanged:
\begin{equation}
    K_{i,j,t} = K_{i,j,t-1}.
\end{equation}

\textbf{(2) Retrieval-Based Response.}
If additional evidence is needed, the Evaluator Agent first formulates a new search query $query_{i,j,t}$ based on the current message and knowledge context, ensuring it does not duplicate any existing query $\{query_{i,j,k}\}^{t-1}_{k=1}$:
\begin{equation}\label{eq:role-play}
    query_{i,j,t} = \mathrm{Evaluator}^m(conf^m, c_{i,j}, M_{i,j,t-1}, K_{i,j,t-1}),
\end{equation}
the agent then invokes an external search tool to retrieve knowledge:
\begin{equation}\label{eq:search}
    k_{i,j,t} = \mathrm{Search}(query_{i,j,t}),
\end{equation}
and the retrieved information is appended to the shared knowledge base:
\begin{equation}
    K_{i,j,t} = \mathrm{Context}(K_{i,j,t-1}, query_{i,j,t}, k_{i,j,t}).
\end{equation}

Then, the agent responds based on the updated knowledge:
\begin{equation}
    [p_{i,j,t}, e_{i,j,t}] = \mathrm{Evaluator}^m(conf^m, c_{i,j}, M_{i,j,t-1}, K_{i,j,t}).
\end{equation}

Finally, the message pool is updated:
\begin{equation}
    M_{i,j,t} = \mathrm{Context}(M_{i,j,t-1}, [p_{i,j,t}, e_{i,j,t}]).
\end{equation}

\textbf{(3) Conditional Retrieval Response.}
The Evaluator Agent first estimates its confidence $c_{i,j,t}$. If $c_{i,j,t} \geq \theta$, it proceeds with a \textbf{direct response} as above. Otherwise, if $c_{i,j,t} < \theta$, it switches to the \textbf{retrieval-based response} workflow, invoking external search to enhance factual grounding before issuing a judgment.

\paragraph{Debate Rules}
\label{para:Debate Rules}

\begin{figure}[!htbp]
    \centering
    \includegraphics[width=1\linewidth]{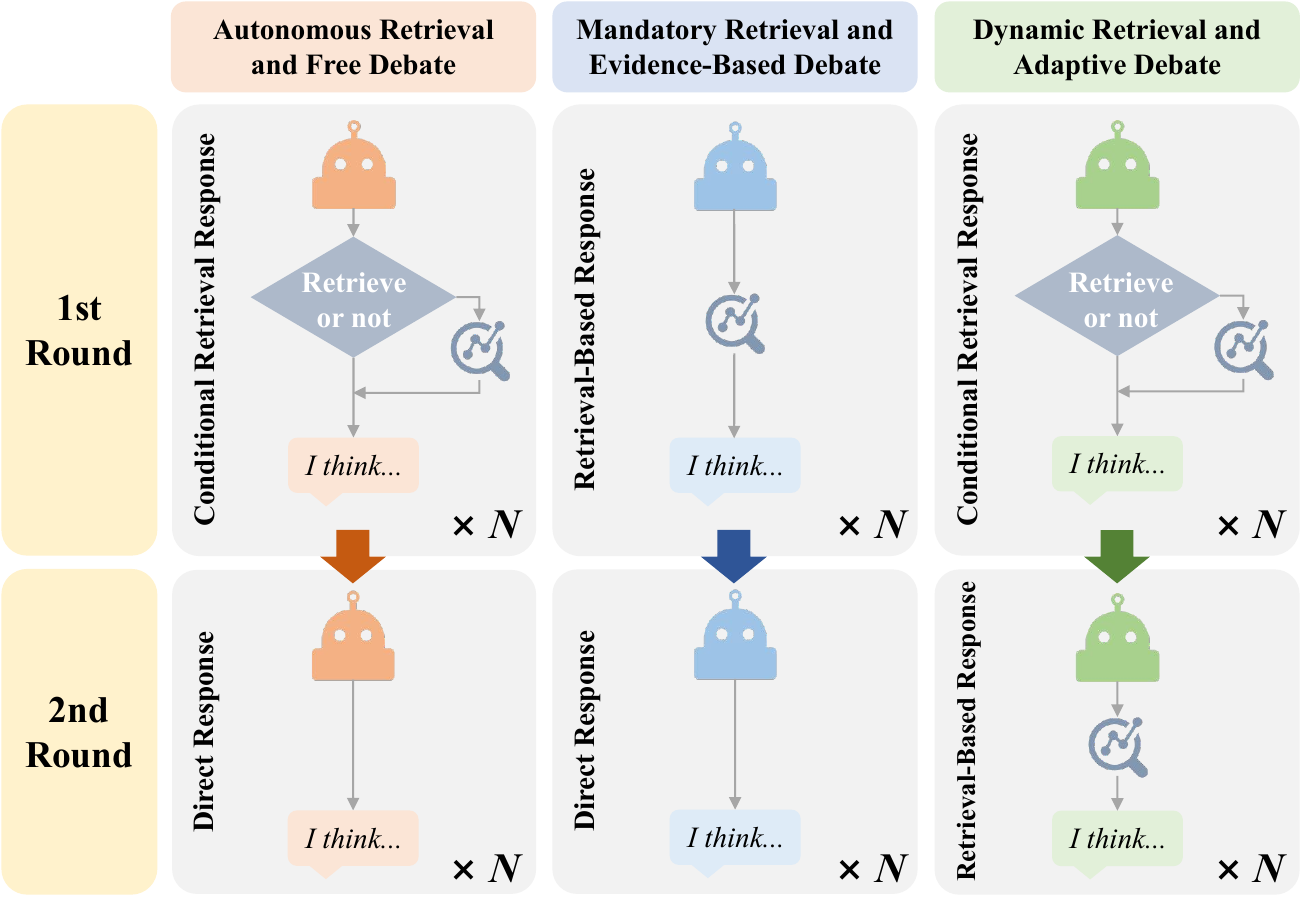}
    \caption{\enspace Three types of debate rules of the MAD-Fact system. For the convenience of systematic analysis, we set the number of debate rounds among multi-agent to 2, and the number of agents \textit{N} to 3.}
    \label{fig:3-debate}
\end{figure}

Based on the three response strategies discussed earlier and inspired by existing works~\cite{Chan2023ChatEvalTB, bo2024reflective}, we further design three types of debate rules to regulate how Evaluator agents interact during the verification process, as illustrated in Figure~\ref{fig:3-debate}. 
These rules enable a systematic exploration of the trade-offs between retrieval cost, prior knowledge utilization, and factual reliability, thus facilitating a comparison of debate strategies within MAD-Fact.

\textbf{Rule 1: Autonomous Retrieval and Free Debate.}  In the first round, each agent may autonomously decide, based on its confidence, whether to invoke external retrieval before speaking, so as to balance knowledge utilization and retrieval cost. In the second round, agents directly deliver their statements while also engaging in discussion with others by drawing on peers’ suggestions or correcting mistakes.  

\textbf{Rule 2: Mandatory Retrieval and Evidence-Based Debate.}  In the first round, retrieval is mandatory before speaking, ensuring that agents compensate for possible knowledge gaps and mitigate overconfidence, a phenomenon where they assign overly high confidence to themselves. In the second round, agents debate with others based on multiple retrieved references, making the discussion more evidence-grounded and reliable.  

\textbf{Rule 3: Dynamic Retrieval and Adaptive Debate.}  In the first round, each agent may autonomously retrieve information before speaking. The debate process in later rounds is then adjusted according to the jury’s consensus: if consensus is reached in the first round, the results are directly output to avoid redundant costs; if not, retrieval becomes mandatory before the second round, so that authoritative references can help resolve conflicts and guide the jury toward agreement.  

\begin{table*}[!t]
    \caption{\enspace Different role descriptions adapted to MAD-Fact.}
    \label{tab:role-play}
    \centering
    \footnotesize
    \setlength{\tabcolsep}{4pt}
    \renewcommand{\arraystretch}{1.2}
    \begin{tabular}{m{0.2\linewidth}m{0.8\linewidth}}
        \hline
        \multicolumn{1}{c}{\textbf{Name}} & \multicolumn{1}{c}{\textbf{Description}}\\
        \hline
        \multicolumn{1}{c}{\textbf{General Public}} & \textit{You are now General Public}, one of the referees in this task. As a member of the general public, you are interested in the claim and eager to get updates on the investigation. You can precisely capture the main meaning of the text rather than fixating on every single word.
        \\   \hline
        \multicolumn{1}{c}{\textbf{Critic}} & \textit{You are Critic}, one of the referees in this task. You are adept at questioning the judgment of others by searching through chains of evidence. In addition, you pay attention to the rigor of the data and are keen to pick up small differences in claims.
        \\   \hline
        \multicolumn{1}{c}{\textbf{News Author}} & \textit{You are News Author}, one of the referees in this task. You focus on the factual basis of the claim and the latest developments, verifying the accuracy of the claim through extensive access to information. When information is insufficient, you tend to search rather than jump to conclusions.
        \\   \hline
        \multicolumn{1}{c}{\textbf{Scientist}} & \textit{You are Scientist}, one of the referees in this task. As a data science research professional, you have a deep background in critical thinking and problem solving skills and are sensitive to data. You are adept at verifying the accuracy of claims by looking at references.
        \\   \hline
        \multicolumn{1}{c}{\textbf{Psychologist}} & \textit{You are Psychologist}, one of the referees in this task. Your job is to study human behavior and mental processes to understand and explain human behavior. Assist others in determining which response is the better one among the available options.
        \\   \hline
        \multicolumn{1}{c}{\textbf{Data Analyst}} & \textit{You are now Data Analyst}, one of the referees in this task.  Specializing in dissecting complex datasets, you approach the claim with a quantitative lens.  You have a knack for gathering relevant data from diverse sources, cleaning and organizing it to extract meaningful insights.
        \\   \hline
    \end{tabular}
\end{table*}

\paragraph{Agent Role-Playing}

In addition, to improve the diversity of the evaluation system, the jury adopts a heterogeneous role assignment mechanism, where each agent is assigned a distinct professional role. This differentiated setup encourages specialized focuses and enables cross-checking from multiple perspectives, thereby reducing factual evaluation biases that may arise from single-perspective blind spots.
Following the setup in ChatEval~\cite{Chan2023ChatEvalTB}, we define six roles for the factuality evaluation task: \textit{Public}, \textit{Critic}, \textit{News Author}, \textit{Scientist}, \textit{Psychologist}, and \textit{Data Analyst}, as shown in Table~\ref{tab:role-play}. 

In addition, at the role description level, we introduce a retrieval incentive mechanism to encourage agents to use external tools, thereby reducing overconfidence and enhancing the reliability of factual verification.

\subsubsection{Judge: Determining the Final Result}
The output opinions and statements from the last round of the jury debate, $\{p^n_{i,j}, e^n_{i,j}\}^N_{n=1}$, are submitted to the Judge agent. The Judge agent aggregates these outputs and determines the final result based on the majority voting principle. This process can be represented as:
\begin{equation}
    p_{i,j} = \mathrm{Judge}(\{p^n_{i,j}, e^n_{i,j}\}^N_{n=1}).
\end{equation}

If the number of agents \textit{N} is even and results in a tie, the output opinion $p^N_{i,j}$ of the last-speaking Evaluator agent is chosen as the final result, since this agent has access to the complete debate process and its output best represents the overall deliberation.  

For a long-form response $a_i$ generated by the evaluated model, along with its corresponding \textit{T} atomic claims $\{c_{i,j}\}^T_{j=1}$, the Judge agent integrates the final evaluation results of each atomic claim  $\{p_{i,j}\}^T_{j=1}$, to calculate the score $s_i$ of the response. This process can be expressed as:
\begin{equation}
    s_i = \mathrm{Calculate}(\{p_{i,j}\}^T_{j=1}).
\end{equation}

The overall score $s$ of the evaluated model on the entire dataset is defined as the arithmetic mean of the scores $s_i$ of all long-form responses, which can be formulated as
$s = \frac{1}{|D|}\sum_{i = 1}^{|D|} s_i$,
where $|D|$ denotes the total number of samples in the dataset. The specific calculation method will be elaborated in the next chapter.

\section{Evaluation Metrics Based on Factual Importance}
\label{sec:metrics}

In this section, we introduce evaluation metrics that account for the varying importance of facts in long-form text. We first describe the construction of a fact importance hierarchy model, which quantifies the relative significance of individual claims, and then present weighted evaluation metrics built upon this model to provide a more nuanced factuality assessment.

\subsection{Fact Importance Hierarchy Model}

\begin{figure*}[!t]
    \centering
    \includegraphics[width=1\textwidth]{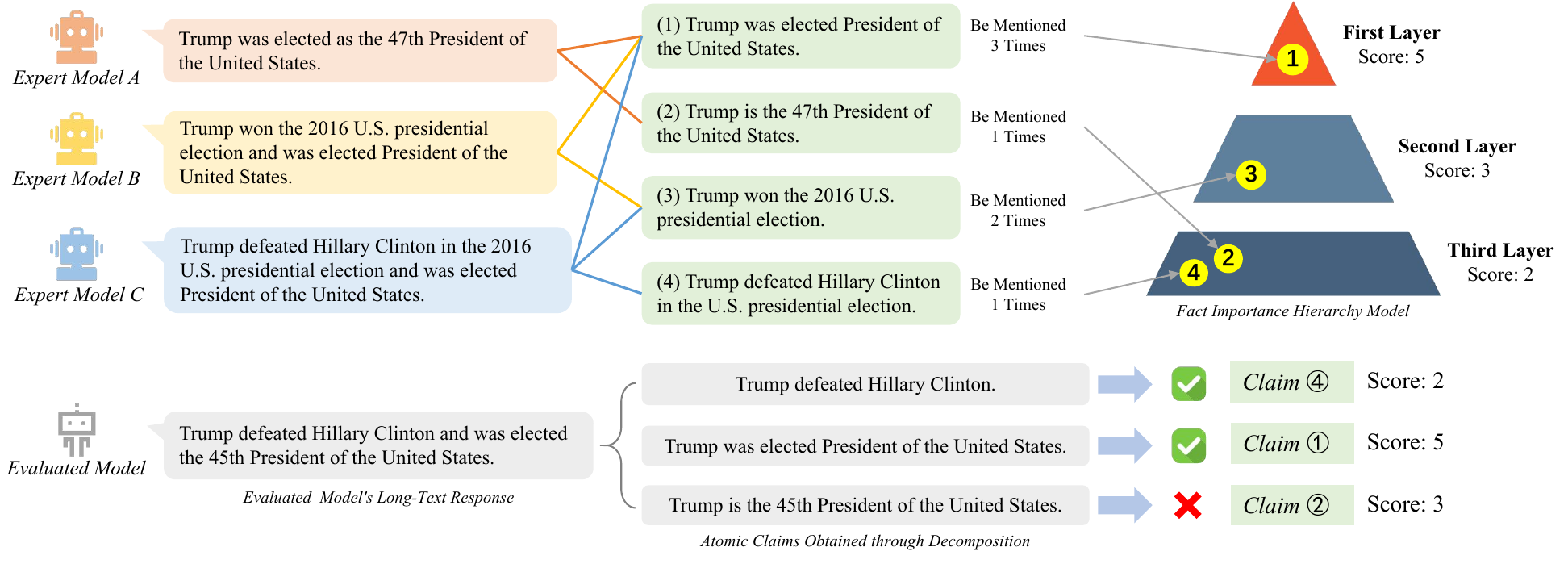}
    \caption{\enspace Overview of the fact importance hierarchy model.}
    \label{fig:jinzita-model}
\end{figure*}

As illustrated in Figure~\ref{fig:jinzita-model}, for a given question $q_i$, we introduce \textit{G} powerful closed-source models $\{Model_j\}^G_{j=1}$ as expert models, each generating a reference answer $\{r_{i,j}\}^G_{j=1}$. This process can be expressed as:
\begin{equation}
    r_{i,j} = \mathrm{Model}_j(q_i).
\end{equation}

Each reference answer is then decomposed into a set of atomic claims $\{[c_{i,j,k}]^{K_j}_{k=1}\}^G_{j=1}$, where the decomposition process can be expressed as:
\begin{equation}
    [c_{i,j,k}]^{K_j}_{k=1} = \mathrm{Clerk}(r_{i,j}),
\end{equation}
with $K_j$ denoting the number of atomic claims in the $j$-th set.  

Next, semantically equivalent atomic claims across different sets are merged to form a single golden set of atomic claims $\{g_{i,k}\}^{K_{g}}_{k=1}$, which is regarded as exhaustive. This merging process can be expressed as:
\begin{equation}
    \{g_{i,k}\}^{K_{gold}}_{k=1} = \mathrm{set}\left(\{[c_{i,j,k}]^{K_j}_{k=1}\}^G_{j=1}\right),
\end{equation}
where $K_{gold}$ denotes the number of atomic claims in the golden set.  

A $G$-level pyramid model $P_{i,G}$ is then constructed, where the levels are ordered from the first layer at the top to the $G$-th layer at the bottom. For each atomic claim in the golden set, we count its frequency \textit{f} of occurrence across the $G$ reference answers, and place it into the $(G-f+1)$-th layer of the pyramid model. The higher the layer an atomic claim belongs to, the more frequently it is mentioned by expert models, and the greater its assigned factual importance weight. The weight of the $m$-th layer in the pyramid model is defined as $\omega_m$, satisfying $\omega_i > \omega_j$ (if  $1 \leq i < j \leq G$).

\subsection{Weighted Evaluation Metrics}
\label{sec:WeightedMetrics}

Given a question $q_i$, we construct the pyramid model $P_{i,G}$ following the aforementioned procedure and obtain the golden set of atomic claims $\{g_{i,k}\}^{K_{gold}}_{k=1}$. For a long-form response $a_i$ to be evaluated, we calculate the weight $\omega_{i,j}$ of each decomposed atomic claim $c_{i,j}$ according to the pyramid model and define the following weighted evaluation metrics.  

\subsubsection{Weighted Precision}
Let $c^{T}_{i,j}$ denote factual atomic claims, with a total count of $|S|$, and $c^{F}_{i,j}$ denote non-factual atomic claims, with a total count of $|N|$. The weighted factual precision score is defined as:
\begin{equation}
    \begin{aligned}
        Prec_w(a_i) =& \frac{\sum_{j=1}^{|S|} \omega_{i,j} \, c^{T}_{i,j}}{\sum_{j=1}^{|S|} \omega_{i,j} \, c^{T}_{i,j} + \sum_{j=1}^{|N|} \omega_{i,j} \, c^{F}_{i,j}} \\
        =& \frac{\sum_{j=1}^{|S|} \omega_{i,j} \, c^{T}_{i,j}}{\sum_{j=1}^{|S|+|N|} \omega_{i,j} \, c_{i,j}}.
    \end{aligned}
\end{equation}

\subsubsection{Weighted Recall}
For a given question $q_i$, the golden set $\{g_{i,k}\}^{K_{gold}}_{k=1}$ is regarded as an exhaustive reference answer. However, responses generated by LLMs often contain redundant information, and the golden set may amplify the error caused by such redundancy. To mitigate this issue, we introduce a hyperparameter $\gamma$ ($\gamma \leq 1$) to account for the gap between the golden set and a perfect answer, thereby alleviating evaluation bias. The weighted factual recall score is defined as:
\begin{equation}
    R_w@\gamma(a_i) = \min\left(\frac{1}{\gamma} \cdot \frac{\sum_{j = 1}^{|S|} \omega_{i,j} \, c^{T}_{i,j}}{\sum_{k = 1}^{K_{gold}} \omega_{i,k} \, g_{i,k}}, ~1\right).
\end{equation}

\subsubsection{Weighted F1-Score}
Based on the weighted precision and weighted recall, the weighted F1-score is defined as:
\begin{equation}
    F_1@\gamma(a_i)=
    \begin{cases}
        \frac{2 \cdot Prec_w(a_i) \cdot R_w@\gamma(a_i)}{Prec_w(a_i)+R_w@\gamma(a_i)}, & \text{if } |S|>0,\\
        0, & \text{if } |S|=0.
    \end{cases}
\end{equation}

\section{Experiments}

Our experiments comprise two parts. In Section~\ref{sec:eval-madfact}, we evaluate the MAD-Fact system on five fact-checking datasets and assess the effects of different debate rules, initialization strategies, and individual system components on its performance. In Section~\ref{sec:eval-llm}, we evaluate nine mainstream LLMs using MAD-Fact (Section~\ref{sec:MAD-Fact}) on LongFact~\cite{wei2024long} and our newly constructed LongHalluQA (Section~\ref{sec:LongHalluQA}), employing weighted evaluation metrics (Section~\ref{sec:metrics}).

\subsection{Evaluation of MAD-Fact}
\label{sec:eval-madfact}

\subsubsection{Datasets}
We evaluate the performance of the MAD-Fact system on four fact-checking datasets: FacTool~\cite{Chern2023FacToolFD}, FELM~\cite{zhao2023felm}, Factcheck-Bench~\cite{Wang2023FactcheckBenchFE}, and BingCheck~\cite{li2023self}. To ensure fair comparison with prior studies, we follow the settings of FIRE~\cite{xie2024fire} and select from the multi-domain data of FacTool and FELM the subsets that require factual knowledge for verification, which we denote as FacToolQA and FELM-WK. To address the class imbalance issue in BingCheck (3,581 atomic claims labeled as TRUE versus 42 labeled as FALSE), we adopt a stratified sampling strategy for balancing. Specifically, we construct the test set by sampling 100 representative examples from the TRUE class while retaining all 42 FALSE examples. To further validate the fact-checking capability of MAD-Fact in the Chinese context, we construct a dataset called FactEval-CN by extracting one representative model statement from each sample in the ChineseFactEval~\cite{wang2023chinesefacteval}, resulting in a total of 125 Chinese factual claims.

All five datasets contain multiple long-form factual QA pairs, along with the corresponding atomic claims decomposed from the long responses and their binary factuality labels. Detailed statistics are shown in Table~\ref{tab:fc-dataset}.

\begin{table}[!htbp]
    \caption{\enspace Statistics of the Fact-Checking Dataset.}
    \label{tab:fc-dataset}
    \centering
    \footnotesize
    \setlength{\tabcolsep}{4pt}
    \renewcommand{\arraystretch}{1.0}
    \begin{tabular}{|p{3cm}|p{3cm}|p{3cm}|p{3cm}|p{3cm}|}
        \toprule
        \multicolumn{1}{c}{\textbf{Dataset}} 
        & \multicolumn{1}{c}{\textbf{Language}}
        & \multicolumn{1}{c}{\textbf{\#TRUE claims}}
        & \multicolumn{1}{c}{\textbf{\#FALSE claims}}
        & \multicolumn{1}{c}{\textbf{All claims}}
        \\
        \midrule
        \multicolumn{1}{c}{FacToolQA} & \multicolumn{1}{c}{English} & \multicolumn{1}{c}{177} & \multicolumn{1}{c}{56} & \multicolumn{1}{c}{233}
        \\   \midrule
        \multicolumn{1}{c}{FELM-WK} & \multicolumn{1}{c}{English} & \multicolumn{1}{c}{99} & \multicolumn{1}{c}{85} & \multicolumn{1}{c}{184}
        \\   \midrule
        \multicolumn{1}{c}{Factcheck-Bench} & \multicolumn{1}{c}{English} & \multicolumn{1}{c}{472} & \multicolumn{1}{c}{159} & \multicolumn{1}{c}{631}
        \\   \midrule
        \multicolumn{1}{c}{BingCheck} & \multicolumn{1}{c}{English} & \multicolumn{1}{c}{100} & \multicolumn{1}{c}{42} & \multicolumn{1}{c}{142}
        \\   \midrule
        \multicolumn{1}{c}{FactEval-CN} & \multicolumn{1}{c}{Chinese} & \multicolumn{1}{c}{70} & \multicolumn{1}{c}{55} & \multicolumn{1}{c}{125}
        \\   \bottomrule
    \end{tabular}
\end{table}

\subsubsection{Baselines}
To ensure a rigorous performance comparison, we select two representative  methods from the fact-checking domain as baselines:

\textbf{(1) SAFE}~\cite{wei2024long}: Utilizes LLM-based agents for long-form factuality evaluation, outperforming crowdsourced human annotators by reaching 72\% agreement and prevailing in 76\% of 100 randomly sampled disagreement cases.

\textbf{(2) FIRE}~\cite{xie2024fire}: Iteratively integrates retrieval and verification, triggering external retrieval only when model confidence falls below a threshold. This design exploits the verifier’s internal knowledge while substantially reducing computational costs without sacrificing accuracy.

\subsubsection{Evaluation Metrics}
We evaluate the system in terms of precision (Prec), recall, and F1 score, separately for the positive class (atomic claims labeled as TRUE) and the negative class (atomic claims labeled as FALSE). Precision refers to the proportion of atomic claims predicted as factual that are indeed labeled TRUE, while recall measures the proportion of TRUE-labeled claims that are correctly identified as factual. Let \textit{TP}, \textit{FP}, \textit{FN}, and \textit{TN} denote true positives, false positives, false negatives, and true negatives, respectively. The evaluation metrics for the positive class are defined as:
\begin{equation}
    \text{Prec} = \frac{TP}{TP + FP},~
    \text{Recall} = \frac{TP}{TP + FN},~
    F_1 = \frac{2 \cdot \text{Prec} \cdot \text{Recall}}{\text{Prec} + \text{Recall}}.
\end{equation}

The evaluation metrics for the negative class are defined analogously.

\subsubsection{Comparative Experiments}

\paragraph{\textit{Q1. How Does MAD-Fact Perform Compared to Baselines?}}

Table~\ref{tab:shiyan-gpt4omini} presents the fact-checking evaluation results of the MAD-Fact system initialized with \texttt{GPT-4o-mini} compared with baseline methods.

\begin{table}[!htbp]
    \caption{\enspace Results of the MAD-Fact system based on \texttt{GPT-4o-mini} and baselines. We set the number of Evaluator agents in the MAD-Fact system to three.}
    \label{tab:shiyan-gpt4omini}
    \centering
    \footnotesize
    \setlength{\tabcolsep}{4pt}
    \renewcommand{\arraystretch}{1.2}
    \begin{tabular}{cccccccccc}
        \toprule
        \multirow{2}{*}{\textbf{Dataset}} & \multirow{2}{*}{\textbf{Method}} & \multicolumn{3}{c}{\textbf{LABEL=True}} & \multicolumn{3}{c}{\textbf{LABEL=False}} \\
        \cmidrule(lr){3-5} \cmidrule(lr){6-8}
        &  & Prec & Recall & F1 & Prec & Recall & F1 \\
        \midrule
        \multirow{3}{*}{\textbf{FactcheckBench}} 
            & SAFE & 0.89 & 0.83 & 0.86 & \textbf{0.70} & 0.79 & \textbf{0.74} \\
            & Fire & 0.91 & \textbf{0.84} & \textbf{0.87} & 0.61 & 0.74 & 0.67 \\
            & MAD-Fact & \textbf{0.94} & 0.77 & 0.84 & 0.55 & \textbf{0.84} & 0.67 \\
        \midrule
        \multirow{3}{*}{\textbf{FacToolQA}} 
            & SAFE & \textbf{0.92} & 0.82 & 0.87 & 0.58 & \textbf{0.79} & \textbf{0.67} \\
            & Fire & 0.87 & \textbf{0.88} & 0.87 & \textbf{0.60} & 0.59 & 0.59 \\
            & MAD-Fact & 0.91 & 0.85 & \textbf{0.88} & \textbf{0.60} & 0.75 & \textbf{0.67} \\
        \midrule
        \multirow{3}{*}{\textbf{BingCheck}} 
            & SAFE & 0.86 & 0.81 & 0.84 & 0.60 & 0.69 & 0.64 \\
            & Fire & 0.87 & \textbf{0.91} & \textbf{0.88} & \textbf{0.74} & 0.67 & 0.70 \\
            & MAD-Fact & \textbf{0.90} & 0.86 & \textbf{0.88} & 0.70 & 0.76 & \textbf{0.73} \\
        \midrule
        \multirow{3}{*}{\textbf{FELM-WK}} 
            & SAFE & 0.61 & 0.76 & 0.68 & 0.61 & 0.44 & 0.51 \\
            & Fire & 0.63 & \textbf{0.82} & 0.71 & 0.67 & 0.44 & 0.53 \\
            & MAD-Fact & \textbf{0.69} & 0.79 & \textbf{0.74} & \textbf{0.70} & \textbf{0.59} & \textbf{0.64} \\
        \midrule
        \multirow{3}{*}{\textbf{FactEval-CN}} 
            & SAFE & 0.85 & \textbf{0.83} & 0.84 & 0.60 & 0.72 & 0.65 \\
            & Fire & 0.86 & 0.82 & 0.74 & 0.63 & 0.73 & 0.68 \\
            & MAD-Fact & \textbf{0.89} & 0.82 & \textbf{0.85} & \textbf{0.64} & \textbf{0.76} & \textbf{0.70} \\
        \bottomrule
    \end{tabular}
\end{table}

In Table~\ref{tab:shiyan-gpt4omini}, the MAD-Fact system adopts the simplest Rule 1 (Autonomous Retrieval and Free Debate). The bold numbers denote the best results for each dataset and label category. The F1-score, which balances both precision and recall, serves as a comprehensive metric for model performance. As shown in Table~\ref{tab:shiyan-gpt4omini}, the MAD-Fact system achieves the best F1-scores in 8 out of 10 comparisons across datasets and label categories, yielding a win rate of 80\%. This demonstrates the superior fact-checking capability of the MAD-Fact system.

\paragraph{\textit{Q2. How Do Debate Rules Influence MAD-Fact's Performance?}}

Further investigation is conducted to explore the impact of different multi-agent debate rules on the performance of the MAD-Fact system. Table~\ref{tab:shiyan-gpt4o} presents the fact-checking evaluation results of the MAD-Fact system initialized with \texttt{GPT-4o}, compared against the baseline methods.

\begin{table}[!htbp]
    \caption{\enspace Results of the MAD-Fact system based on \texttt{GPT-4o} and baselines. The three debate rules introduced in Section \ref{para:Debate Rules} are denoted as follows: Rule 1 (Autonomous Retrieval and Free Debate) is referred to as MAD-Fact-\textit{free}; Rule 2 (Mandatory Retrieval and Evidence-Based Debate) is referred to as MAD-Fact-\textit{search}; and Rule 3 (Dynamic Retrieval and Adaptive Debate) is referred to as MAD-Fact-\textit{adapt}. }
    \label{tab:shiyan-gpt4o}
    \centering
    \footnotesize
    \setlength{\tabcolsep}{4pt}
    \renewcommand{\arraystretch}{1.2}
    \begin{tabular}{cccccccccc}
        \toprule
        \multirow{2}{*}{\textbf{Dataset}} & \multirow{2}{*}{\textbf{Method}} & \multicolumn{3}{c}{\textbf{LABEL=True}} & \multicolumn{3}{c}{\textbf{LABEL=False}} \\
        \cmidrule(lr){3-5} \cmidrule(lr){6-8}
        &  & Prec & Recall & F1 & Prec & Recall & F1 \\
        \midrule
        \multirow{5}{*}{\textbf{FactcheckBench}} 
            & SAFE & \textbf{0.94} & 0.74 & 0.83 & 0.62 & \textbf{0.90} & \textbf{0.74} \\
            & Fire & 0.92 & \textbf{0.79} & \textbf{0.85} & 0.56 & 0.79 & 0.66 \\
            & MAD-Fact-\textit{free} & 0.92 & 0.76 & 0.83 & 0.53 & 0.80 & 0.64 \\
            & MAD-Fact-\textit{search} & 0.92 & \textbf{0.79} & \textbf{0.85} & \textbf{0.65} & 0.87 & \textbf{0.74} \\
            & MAD-Fact-\textit{adapt} & 0.91 & 0.72 & 0.81 & 0.60 & 0.85 & 0.70 \\
        \midrule
        \multirow{5}{*}{\textbf{FacToolQA}} 
            & SAFE & \textbf{0.92} & \textbf{0.88} & \textbf{0.90} & \textbf{0.66} & \textbf{0.77} & \textbf{0.71} \\
            & Fire & \textbf{0.92} & \textbf{0.88} & \textbf{0.90} & 0.65 & 0.71 & 0.68 \\
            & MAD-Fact-\textit{free} & 0.90 & \textbf{0.88} & 0.89 & 0.64 & 0.68 & 0.66 \\
            & MAD-Fact-\textit{search} & 0.91 & \textbf{0.88} & \textbf{0.90} & \textbf{0.66} & 0.73 & 0.69 \\
            & MAD-Fact-\textit{adapt} & 0.91 & 0.85 & 0.88 & 0.61 & 0.73 & 0.67 \\
        \midrule
        \multirow{5}{*}{\textbf{BingCheck}} 
            & SAFE & 0.86 & 0.81 & 0.84 & 0.71 & 0.60 & 0.65 \\
            & Fire & 0.86 & 0.88 & 0.87 & 0.70 & 0.67 & 0.68 \\
            & MAD-Fact-\textit{free} & \textbf{0.87} & 0.90 & \textbf{0.89} & 0.74 & \textbf{0.69} & \textbf{0.72} \\
            & MAD-Fact-\textit{search} & 0.85 & 0.88 & 0.87 & 0.69 & 0.64 & 0.67 \\
            & MAD-Fact-\textit{adapt} & \textbf{0.87} & \textbf{0.92} & \textbf{0.89} & \textbf{0.78} & 0.67 & \textbf{0.72} \\
        \midrule
        \multirow{5}{*}{\textbf{FELM-WK}} 
            & SAFE & 0.70 & 0.80 & 0.75 & 0.72 & \textbf{0.60} & 0.65 \\
            & Fire & 0.70 & 0.86 & 0.77 & 0.77 & 0.54 & 0.63 \\
            & MAD-Fact-\textit{free} & \textbf{0.72} & \textbf{0.89} & \textbf{0.79} & \textbf{0.82} & 0.59 & \textbf{0.68} \\
            & MAD-Fact-\textit{search} & 0.70 & 0.86 & 0.77 & 0.77 & 0.56 & 0.65 \\
            & MAD-Fact-\textit{adapt} & 0.70 & 0.84 & 0.76 & 0.76 & 0.59 & 0.66 \\
        \midrule
        \multirow{5}{*}{\textbf{FactEval-CN}} 
            & SAFE & 0.85 & 0.85 & 0.85 & 0.64 & 0.72 & 0.68 \\
            & Fire & 0.86 & 0.83 & 0.84 & 0.62 & 0.70 & 0.66 \\
            & MAD-Fact-\textit{free} & \textbf{0.90} & 0.83 & \textbf{0.86} & \textbf{0.66} & \textbf{0.76} & \textbf{0.71} \\
            & MAD-Fact-\textit{search} & 0.87 & 0.83 & 0.85 & 0.64 & 0.71 & 0.67 \\
            & MAD-Fact-\textit{adapt} & 0.81 & \textbf{0.87} & 0.84 & 0.64 & 0.64 & 0.64 \\
        \bottomrule
    \end{tabular}
\end{table}

In Table~\ref{tab:shiyan-gpt4o}, the bold numbers indicate the best results for each dataset and label category. The experimental results show that the MAD-Fact system employing Rule 1 and Rule 2 demonstrates superior fact-checking performance, validating the importance of synergizing external retrieval tools with the model’s internal knowledge base. In contrast, the system based on Rule 3 performs less effectively. Error analysis reveals that its weakness mainly stems from erroneous consensus reached in the first round of debate, which, due to the lack of subsequent correction opportunities, becomes entrenched. Specifically, when the system arrives at an incorrect conclusion in the early stage, the dynamic termination mechanism prematurely halts the debate process, depriving the multi-agent system of the opportunity for deeper reasoning through multi-round interactions.

\paragraph{\textit{Q3. Does Initializing with Different Model Families Affect MAD-Fact?}}

Further exploration was conducted to investigate the impact of initializing the MAD-Fact system with models from different families. Specifically, we initialized the three Evaluator agents in the MAD-Fact system with \texttt{GPT-4o-mini}, \texttt{DeepSeek-V3}, and \texttt{Claude-3-Haiku}, denoted as MAD-Fact-\textit{various}, and compared it with the MAD-Fact system initialized solely with \texttt{GPT-4o-mini}. Using the simplest Rule 1 (Autonomous Retrieval and Free Debate), Table~\ref{tab:shiyan-various} presents the fact-checking evaluation results of the MAD-Fact system based on multi-model initialization.

\begin{table}[!htbp]
    \caption{\enspace Results of the MAD-Fact system based on multi-model initialization.}
    \label{tab:shiyan-various}
    \centering
    \footnotesize
    \setlength{\tabcolsep}{4pt}
    \renewcommand{\arraystretch}{1.2}
    \begin{tabular}{cccccccccc}
        \toprule
        \multirow{2}{*}{\textbf{Dataset}} & \multirow{2}{*}{\textbf{Method}} & \multicolumn{3}{c}{\textbf{LABEL=True}} & \multicolumn{3}{c}{\textbf{LABEL=False}} \\
        \cmidrule(lr){3-5} \cmidrule(lr){6-8}
        &  & Prec & Recall & F1 & Prec & Recall & F1 \\
        \midrule
        \multirow{2}{*}{\textbf{FactcheckBench}} 
            & MAD-Fact                  & 0.94 & 0.77 & 0.84 & 0.55 & 0.84 & 0.67 \\
            & MAD-Fact-\textit{various} & 0.92 & 0.78 & 0.84 & 0.55 & 0.79 & 0.64 \\
        \midrule
        \multirow{2}{*}{\textbf{FacToolQA}} 
            & MAD-Fact                  & 0.91 & 0.85 & 0.88 & 0.60 & 0.73 & 0.66 \\
            & MAD-Fact-\textit{various} & 0.89 & 0.87 & 0.88 & 0.62 & 0.66 & 0.64 \\
        \midrule
        \multirow{2}{*}{\textbf{BingCheck}} 
            & MAD-Fact                  & 0.90 & 0.86 & 0.88 & 0.70 & 0.76 & 0.73 \\
            & MAD-Fact-\textit{various} & 0.89 & 0.87 & 0.88 & 0.70 & 0.74 & 0.72 \\
        \midrule
        \multirow{2}{*}{\textbf{FELM-WK}} 
            & MAD-Fact                  & 0.69 & 0.79 & 0.74 & 0.70 & 0.59 & 0.64 \\
            & MAD-Fact-\textit{various} & 0.67 & 0.83 & 0.74 & 0.73 & 0.53 & 0.61 \\
        \bottomrule
    \end{tabular}
\end{table}

The experimental results indicate that the MAD-Fact system initialized with multiple models did not achieve significant performance improvement; instead, it showed a slight decline compared to the system initialized with a single model. A deeper analysis of the misclassified cases reveals that, relative to the single-model system, the agents in the multi-model initialization setting exhibited a markedly higher frequency of misleading each other, and reaching consensus within the predefined number of debate rounds became more difficult. Based on these findings, subsequent evaluations adopt the MAD-Fact system initialized with a single model.

\subsubsection{Ablation Study}
To validate the effectiveness of each module in the multi-agent debate system, we designed a series of ablation experiments. The multi-agent debate system was examined in the following three variants:

\textbf{(1)w/o Role-Playing}: Removing the diversified role configurations (i.e., modifying the agent profile $conf^m$ in Equation~\eqref{eq:role-play}), such that all Evaluator agents have exactly the same role settings;

\textbf{(2)w/o Debate}: Removing the multi-round debate process, where Evaluator agents deliver only one round of statements before the Evaluator agent aggregates the results;

\textbf{(3)w/o Search}: Removing the retrieval module (i.e., Equation~\eqref{eq:search}), prohibiting Evaluator agents from invoking external retrieval tools to access reference materials before their statements.

Table~\ref{tab:xiaorong-shiyan} presents the results of the ablation study based on the MAD-Fact system initialized with the gpt-4o-mini model. The experimental results demonstrate that each module in the multi-agent debate system contributes to the overall performance.

\begin{table}[!htbp]
    \caption{\enspace Results of the ablation experiment of the MAD-Fact system. We set the number of Evaluator agents to 3 and applied the simplest Rule 1.}
    \label{tab:xiaorong-shiyan}
    \centering
    \footnotesize
    \setlength{\tabcolsep}{4pt}
    \renewcommand{\arraystretch}{1.2}
    \begin{tabular}{clcccccccc}
        \toprule
        \multirow{2}{*}{\textbf{Dataset}} & \multirow{2}{*}{\textbf{Method}} & \multicolumn{3}{c}{\textbf{LABEL=True}} & \multicolumn{3}{c}{\textbf{LABEL=False}} \\
        \cmidrule(lr){3-5} \cmidrule(lr){6-8}
        &  & Prec & Recall & F1 & Prec & Recall & F1 \\
        \midrule
        \multirow{4}{*}{\textbf{BingCheck}} 
            & MAD-Fact         & 0.90 & 0.86 & 0.88 & 0.70 & 0.76 & 0.73  \\
            & w/o Role-Playing & 0.89 & 0.87 & 0.88 & 0.70 & 0.74 & 0.72 \\
            & w/o Debate    & 0.87 & 0.78 & 0.82 & 0.58 & 0.71 & 0.64 \\
            & w/o Search    & 0.85 & 0.83 & 0.84 & 0.63 & 0.52 & 0.57 \\
        \midrule
        \multirow{4}{*}{\textbf{FacToolQA}} 
            & MAD-Fact         & 0.91 & 0.85 & 0.88 & 0.60 & 0.75 & 0.67 \\
            & w/o Role-Playing & 0.91 & 0.83 & 0.87 & 0.58 & 0.75 & 0.66 \\
            & w/o Debate    & 0.89 & 0.85 & 0.87 & 0.57 & 0.66 & 0.61 \\
            & w/o Search    & 0.84 & 0.84 & 0.84 & 0.48 & 0.48 & 0.48 \\
        \bottomrule
    \end{tabular}
\end{table}

\subsubsection{Case Study}

During the experiments, we observed several exciting cases. In debate scenarios, multi-agent systems exhibited many human-like traits: they were able to adhere to their own viewpoints, actively correct their peers’ mistakes, and continuously engage in self-reflection. Such human-like behaviors may be one of the key factors contributing to the superior performance of multi-agent systems compared to single-agent systems. Below, we present several interesting results we observed:

\begin{figure*}[!t]
    \centering
    \includegraphics[width=1\linewidth]{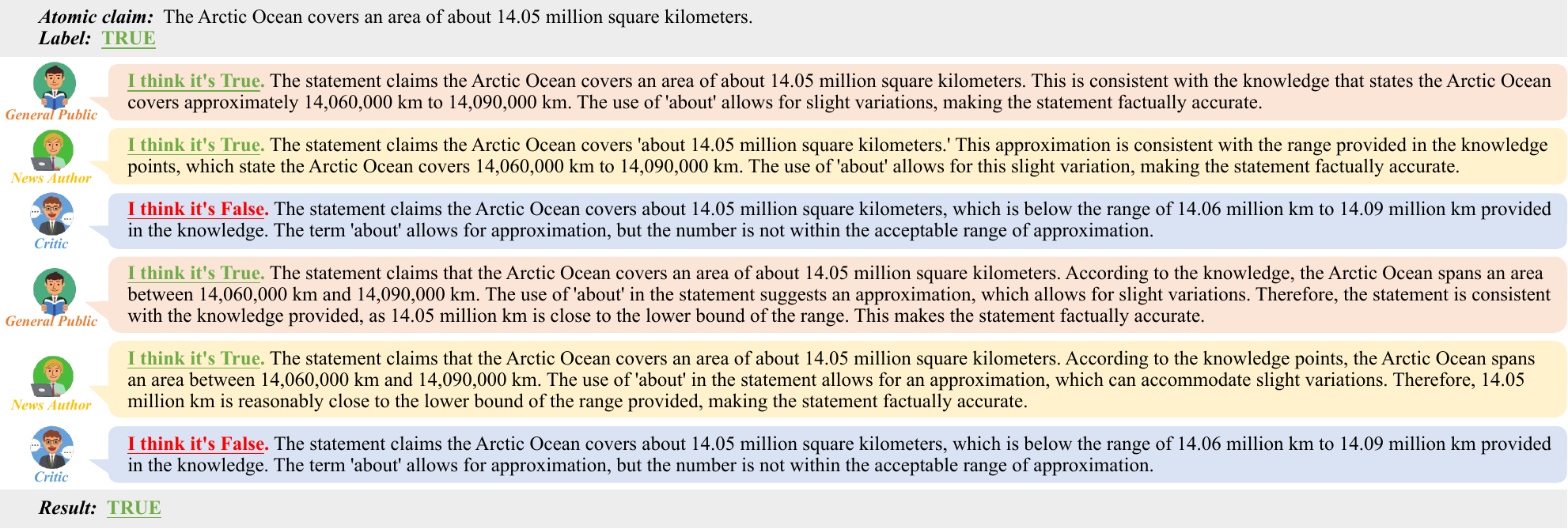}
    \caption{\enspace Case 1 of multi-agent debate: agents adhere to their own views.}
    \label{fig:example-1}
\end{figure*}

\textbf{(1) Agents do not easily change their stance when facing differing opinions, but instead adhere to their own viewpoints.} As shown in Figure~\ref{fig:example-1}: in the first round of debate, \textit{Public} and \textit{News Author} provided the correct answers, while \textit{Critic} gave an incorrect one. In the second round, \textit{Public} and \textit{News Author} did not readily accept the view of the \textit{Critic} but instead held to their own judgments. Eventually, during the voting process, the multi-agent system reached the correct conclusion.

\begin{figure*}[!t]
    \centering
    \includegraphics[width=1\linewidth]{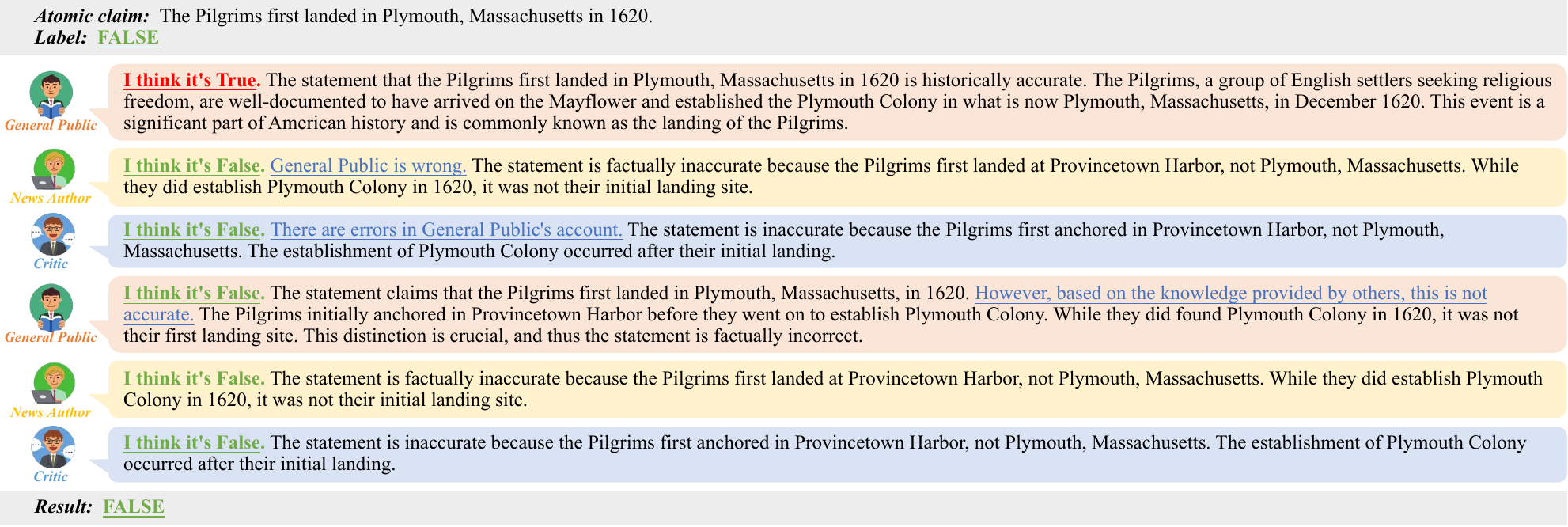}
    \caption{\enspace Case 2 of multi-agent debate: agents correct their peers' mistakes.}
    \label{fig:example-2}
\end{figure*}

\textbf{(2) Agents actively correct their peers' mistakes when they detect problems in their statements.} As shown in Figure~\ref{fig:example-2}: in the first round of debate, \textit{Public} gave an incorrect answer, while \textit{News Author} and \textit{Critic} provided the correct answers and attempted to correct \textit{Public}, At the beginning of the second round, with the help of its peers, \textit{Public} realized its mistake, revised its position, and gave the correct answer, thereby reaching consensus with \textit{News Author} and \textit{Critic}.

\begin{figure*}[!t]
    \centering
    \includegraphics[width=1\linewidth]{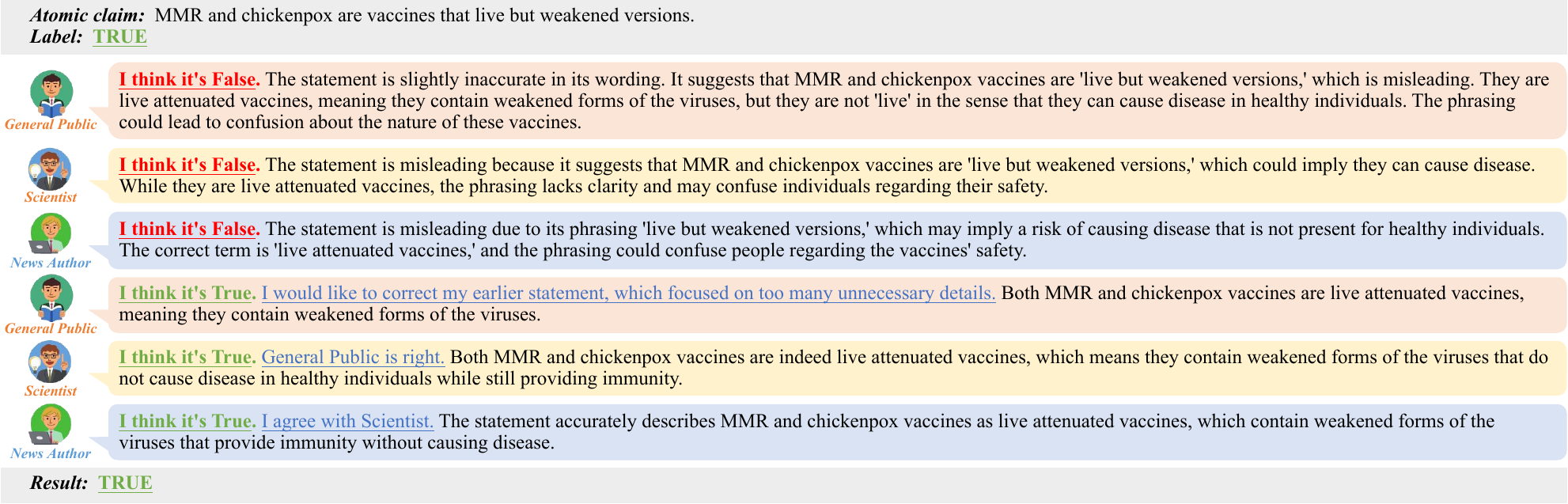}
    \caption{\enspace Case 3 of multi-agent debate: agents continuously conduct self-reflection.}
    \label{fig:example-3}
\end{figure*}

\textbf{(3) Even when consensus is achieved among peers, agents continue to engage in self-reflection.} As illustrated in Figure~\ref{fig:example-3}: in the first round of debate, \textit{Public}, \textit{Scientist}, and \textit{News Author} all gave incorrect answers, reaching a wrong consensus. However, \textit{Public} continuously reflected on its viewpoint and, in the second round, changed its stance, providing the correct answer. This also introduced a new explanation to \textit{Scientist} and \textit{News Author}, prompting them to engage in reflection as well. Eventually, the group reached the correct consensus.

\subsection{Factuality Evaluation Results of LLMs}
\label{sec:eval-llm}

\subsubsection{Datasets}
This study employs two long-form factuality evaluation datasets: LongFact~\cite{wei2024long} and LongHalluQA. LongFact is the first systematically constructed cross-domain benchmark for long-form factuality evaluation, encompassing 38 manually curated topics and 2,280 factual questions, where models are required to generate long-form textual answers. Following the setup of SAFE~\cite{wei2024long}, we apply stratified sampling to select 250 instances from LongFact with a balanced topic distribution as our test set. LongHalluQA, newly introduced in this paper, is a Chinese benchmark for long-form factuality evaluation, with details provided in Section~\ref{sec:LongHalluQA}. From this dataset, we sampled 250 high-quality instances for evaluation.

\subsubsection{Evaluated Models}
In this study, we carefully select 9 representative models from 7 different families, both domestic and international, to conduct a comprehensive and in-depth evaluation:

\begin{itemize}
    \item \textbf{Closed-source models:}
        Claude series: \texttt{Claude-3-Sonnet};
        OpenAI series: \texttt{GPT-3.5-Turbo}, \texttt{GPT-4-Turbo};
        Doubao series: \texttt{Doubao-1.5-Pro-32k};
        Kimi series: \texttt{Moonshot-V1-8k}.

    \item \textbf{Open-source models:}
        DeepSeek series: \texttt{DeepSeek-V3};
        Llama series: \texttt{Llama3.1-70B};
        Qwen series: \texttt{Qwen2.5-72B-Instruct}, \texttt{QwQ-32B}.
        
\end{itemize}

\subsubsection{Evaluation Setup}
In the MAD-Fact system, the Clerk and Evaluator agents are initialized based on \texttt{GPT-4o-mini}, while the Judge agent is initialized based on \texttt{Llama3.3-70B-instruct}. The number of Evaluator agents is set to 3, and the debate protocol follows Rule 1: Autonomous Retrieval and Free Debate. 

We adopt the weighted evaluation metrics introduced in section \ref{sec:WeightedMetrics}. Among them, \texttt{GPT-4o}, \texttt{DeepSeek-R1}, and \texttt{Claude-3-Opus} are selected as expert models due to their comprehensive and strong performance. The factual importance score at the $k$-th level of the pyramid model is defined as $\omega=5-k$. Specifically, key points mentioned by all three expert models receive a score of 4, those mentioned by two expert models receive a score of 3, and so forth.

\subsubsection{Results Analysis}
\begin{figure}[!htbp]
    \centering
    \includegraphics[width=1\linewidth]{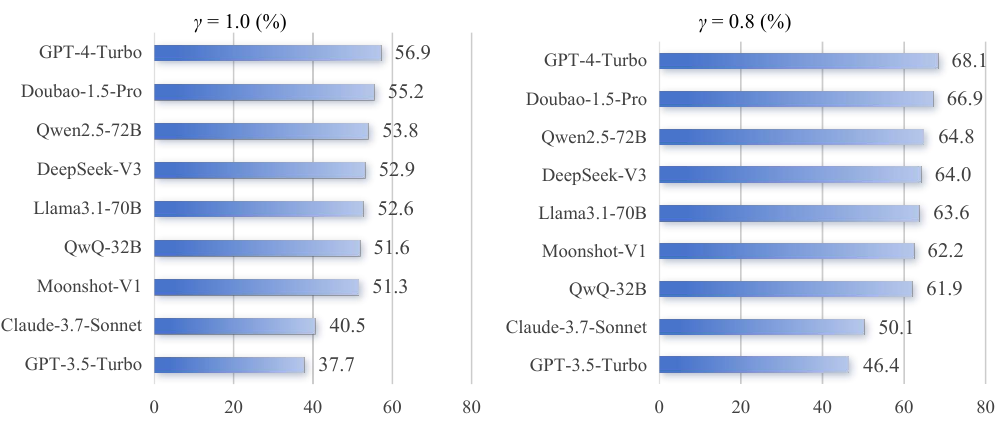}
    \caption{\enspace The evaluation performance of the selected models on LongFact.}
    \label{fig:longfact}
\end{figure}

Figure~\ref{fig:longfact} illustrates the evaluation performance of the selected models on the LongFact dataset, ranked in descending order of $F_1@\gamma$. $F_1@\gamma$ is calculated under $\gamma=1.0$ and $\gamma=0.8$, and the model rankings remain relatively stable across different $\gamma$ values. Detailed results are shown in Table~\ref{tab:longfact-shiyan}.

\begin{table}[!htbp]
    \caption{\enspace The detailed evaluation results of the selected models on LongFact. Bold numbers indicate the best results and underlined numbers denote the second-best results.}
    \label{tab:longfact-shiyan}
    \centering
    \footnotesize
    \setlength{\tabcolsep}{4pt}
    \renewcommand{\arraystretch}{1.2}
    \begin{tabular}{ccccccccc}
        \toprule
        \multirow{2}{*}{\textbf{Dataset}} 
        & \multirow{2}{*}{\textbf{Model}} 
        & \multirow{2}{*}{Precision}
        & \multicolumn{2}{c}{\textbf{$\gamma=1.0$}} 
        & \multicolumn{2}{c}{\textbf{$\gamma=0.8$}} \\
        \cmidrule(lr){4-5} 
        \cmidrule(lr){6-7}
        & & & Recall & F1 & Recall & F1 \\
        \midrule
        \multirow{9}{*}{\textbf{LongFact}} 
            & GPT-4-Turbo           & \underline{0.877} & \textbf{0.445} & \textbf{0.569} & \textbf{0.557} & \textbf{0.681}  \\
            & Doubao-1.5-Pro        & 0.867 & \underline{0.436} & \underline{0.552} & \underline{0.545} & \underline{0.669}  \\
            & Qwen2.5-72B           & 0.847 & 0.420 & 0.538 & 0.525 & 0.648  \\
            & DeepSeek-V3           & 0.838 & 0.414 & 0.529 & 0.518 & 0.640  \\
            & Llama3.1-70B          & 0.819 & 0.416 & 0.526 & 0.520 & 0.636  \\
            & QwQ-32B               & 0.763 & 0.416 & 0.516 & 0.520 & 0.619  \\
            & Moonshot-V1           & 0.837 & 0.396 & 0.513 & 0.495 & 0.622  \\
            & Claude-3.7-Sonnet     & 0.874 & 0.281 & 0.405 & 0.351 & 0.501  \\
            & GPT-3.5-Turbo         & \textbf{0.908} & 0.249 & 0.377 & 0.312 & 0.464  \\
        \bottomrule
    \end{tabular}
\end{table}

As shown in Figure~\ref{fig:longfact} and Table~\ref{tab:longfact-shiyan}, models with larger parameter sizes and more recent releases generally demonstrate better factual consistency in long-form scenarios. For instance, \texttt{GPT-4-Turbo} outperforms \texttt{GPT-3.5-Turbo}. Under the two selected $\gamma$ values, the best-performing model in terms of factuality is \texttt{GPT-4-Turbo}, consistent with the findings reported by SAFE~\cite{wei2024long}. The performance gap between the domestic model \texttt{Doubao-1.5-Pro} and \texttt{GPT-4-Turbo} is minimal, only about 1–2 percentage points, highlighting the strong potential and advantages of domestic models. \texttt{GPT-3.5-Turbo}, as a relatively older model, is surpassed by most newer models. It is noteworthy that \texttt{Claude-3.7-Sonnet} and \texttt{QwQ-32B}, both widely regarded as strong and recently released models, did not achieve high rankings. This is primarily because \texttt{Claude-3.7-Sonnet} has a relatively low factual recall score (0.501 under $\gamma=0.8$, ranking 8th among the selected models), while \texttt{QwQ-32B} has a relatively low factual precision score (0.763, ranking 9th). These results indicate that strong reasoning or coding capabilities do not necessarily guarantee superior factuality.

\begin{figure}[!htbp]
    \centering
    \includegraphics[width=1\linewidth]{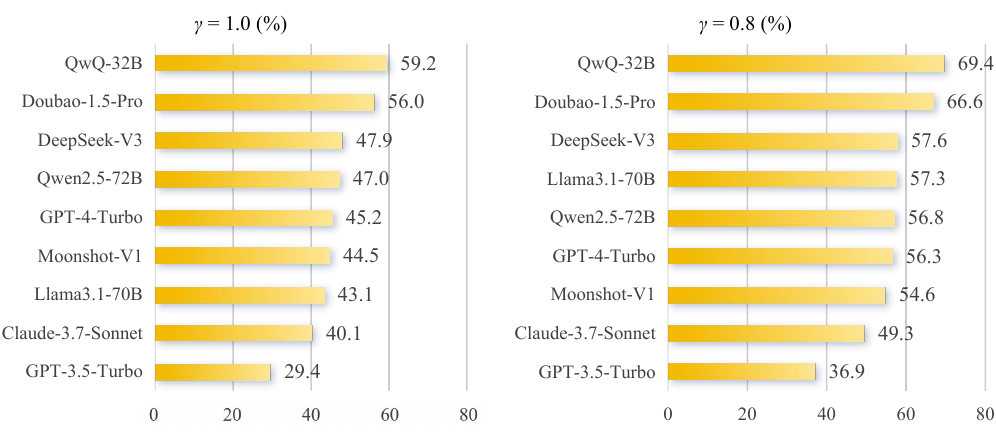}
    \caption{\enspace The evaluation performance of the selected models on LongHalluQA.}
    \label{fig:longhalluqa}
\end{figure}

Figure~\ref{fig:longhalluqa} presents the evaluation performance of the selected models on the LongHalluQA dataset, ranked in descending order of $F_1@\gamma$. $F_1@\gamma$ is calculated under $\gamma=1.0$ and $\gamma=0.8$, and the model rankings remain largely stable, with only slight variations in the middle range. Detailed results are provided in Table~\ref{tab:longhalluqa-shiyan}.

As shown in Figure~\ref{fig:longhalluqa} and Table~\ref{tab:longhalluqa-shiyan}, domestic models exhibit notable advantages in factuality evaluation for Chinese long-form responses. Under the two selected $\gamma$ values, the top three models in terms of factuality are \texttt{QwQ-32B}, \texttt{Doubao-1.5-Pro}, and \texttt{DeepSeek-V3}, all domestic models. Among them, \texttt{QwQ-32B} and \texttt{Doubao-1.5-Pro} clearly outperform the others, exceeding the third-ranked \texttt{DeepSeek-V3} by nearly 10 percentage points. In contrast, \texttt{GPT-4-Turbo} performs worse on the Chinese dataset than on the English dataset, ranking 5th among the selected models when $\gamma=1.0$ and 6th when $\gamma=0.8$, indicating a notable issue of cultural bias~\cite{xu2024self} and an imbalance in multilingual capabilities.

\begin{table}[!htbp]
    \caption{\enspace The detailed evaluation results of the selected models on LongHalluQA. Bold numbers indicate the best results and underlined numbers denote the second-best results.}
    \label{tab:longhalluqa-shiyan}
    \centering
    \footnotesize
    \setlength{\tabcolsep}{4pt}
    \renewcommand{\arraystretch}{1.2}
    \begin{tabular}{ccccccccc}
        \toprule
        \multirow{2}{*}{\textbf{Dataset}} 
        & \multirow{2}{*}{\textbf{Model}} 
        & \multirow{2}{*}{Precision}
        & \multicolumn{2}{c}{\textbf{$\gamma=1.0$}} 
        & \multicolumn{2}{c}{\textbf{$\gamma=0.8$}} \\
        \cmidrule(lr){4-5} 
        \cmidrule(lr){6-7}
        & & & Recall & F1 & Recall & F1 \\
        \midrule
        \multirow{9}{*}{\textbf{LongHalluQA}} 
            & QwQ-32B            & 0.759 & \textbf{0.511} & \textbf{0.592} & \textbf{0.639} & \textbf{0.694}  \\
            & Doubao-1.5-Pro     & \textbf{0.823} & \underline{0.447} & \underline{0.560} & \underline{0.559} & \underline{0.666}  \\
            & DeepSeek-V3        & 0.773 & 0.367 & 0.479 & 0.459 & 0.576  \\
            & Qwen2.5-72B        & 0.807 & 0.351 & 0.470 & 0.439 & 0.568  \\
            & GPT-4-Turbo        & 0.806 & 0.346 & 0.452 & 0.433 & 0.563  \\
            & Moonshot-V1        & \underline{0.819} & 0.328 & 0.445 & 0.410 & 0.546  \\
            & Llama3.1-70B       & 0.787 & 0.360 & 0.431 & 0.450 & 0.573  \\
            & Claude-3.7-Sonnet  & 0.763 & 0.291 & 0.401 & 0.364 & 0.493  \\
            & GPT-3.5-Turbo      & 0.758 & 0.195 & 0.294 & 0.244 & 0.369  \\
        \bottomrule
    \end{tabular}
\end{table}

\begin{figure}[!htbp]
    \centering
    \includegraphics[width=0.5\linewidth]{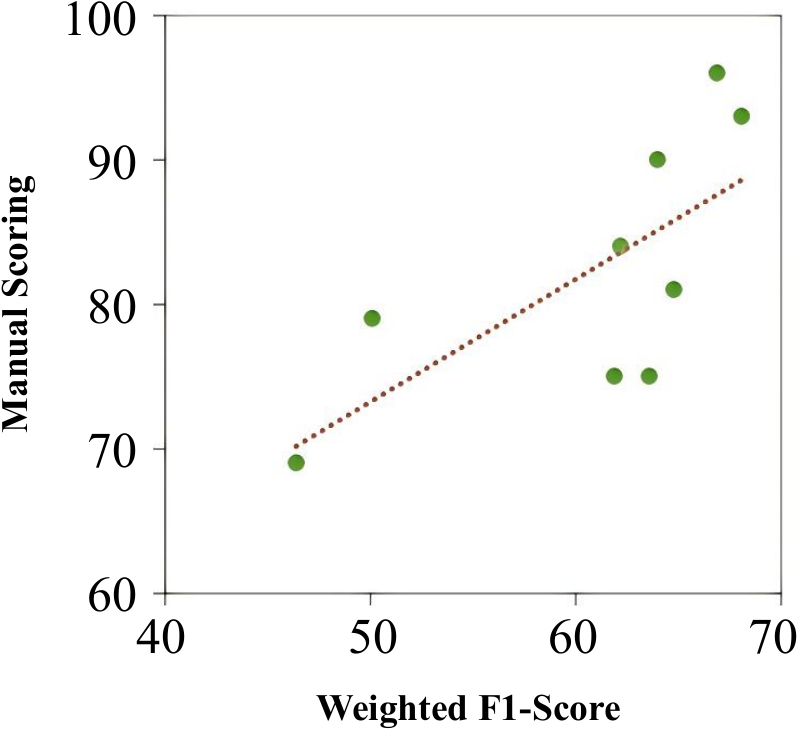} 
    \caption{Correlation analysis between weighted evaluation metrics and human ratings.}
    \label{fig:pearson}
\end{figure}

To verify the effectiveness of the weighted evaluation metric based on the hierarchical importance of facts, we established a review panel consisting of two individuals with undergraduate degrees. From the LongFact dataset, 50 questions were sampled, and the outputs of different models were manually rated. The correlation between the human ratings and the weighted $F_1$ score ($\gamma=0.8$) was then calculated, as shown in Figure~\ref{fig:pearson}. The results indicate that the Pearson correlation coefficient reached a statistically significant value of $r = 0.701$ ($p = 0.036$), demonstrating that the proposed weighted evaluation metric aligns well with human judgments in assessing the factuality of long-form responses.

\section{Discussion}

This paper presents a systematic study on long-form factuality evaluation by LLMs. Nonetheless, limitations remain in dataset representativeness, multi-agent efficiency, and communication reliability. Future research can explore the following directions:

\textbf{Expanding and grounding factuality evaluation datasets.} While LongHalluQA extends existing short-text datasets to support long-form evaluation, the data are still derived from curated sources rather than real-world user-generated content. Future research should incorporate diverse, high-risk domains such as biomedical, finance, and law, and integrate practical real-world sources to improve coverage, generalization, and applicability of long-form factuality evaluation.

\textbf{Optimizing the cost-effectiveness trade-off in multi-agent systems.} The MAD-Fact system enhances fact-checking performance through multi-agent debate, but its reliance on multiple models increases token consumption and operational costs. Future work could explore lightweight collaboration strategies, such as dynamically adjusting the number of agents, optimizing debate rounds, or employing model distillation and parameter sharing to reduce computational overhead. Integrating reinforcement learning to select optimal collaboration modes based on context could further improve efficiency without sacrificing evaluation accuracy, supporting deployment from experimental settings to real-world high-risk scenarios.

\textbf{Mitigating communication hallucinations in multi-agent systems.} During multi-agent debates, agents may reach misleading consensus due to confidently incorrect responses, causing erroneous conclusions. Yoffe et al.~\cite{yoffe2024debunc} attribute this issue to communication failures among agents, referred to as communication hallucinations. Future research can focus on: (i) confidence-based weighting of agent responses to downweight unreliable viewpoints; (ii) cross-agent consistency verification using logical reasoning to detect contradictions; (iii) adversarial training that injects incorrect viewpoints to test and enhance system robustness. Progress in these areas would strengthen the reliability of multi-agent systems in complex long-form evaluation tasks.

\section{Conclusion}

We present a unified framework for long-form factuality evaluation of LLMs, combining large-scale benchmarks, multi-agent verification, and weighted evaluation metrics. We introduce LongHalluQA, a Chinese long-form factuality dataset, and MAD-Fact, a multi-agent debate system designed to mitigate single-model bias. Additionally, we propose hierarchical modeling of factual importance to guide weighted metrics that closely reflect human judgments. Extensive experiments demonstrate MAD-Fact's effectiveness, with evaluations on LongFact and LongHalluQA showing that larger models generally outperform smaller ones, while domestic models excel on Chinese-language tasks. We believe this paper can serve as a foundation for future research on long-form factuality evaluation and guide the development of more reliable LLMs.

\section*{Acknowledgement}
This work is supported by the National Natural Science Foundation of China (No.62402491) and the China Postdoctoral Science Foundation (No.2025M771524).

\bibliographystyle{fcs}
\bibliography{ref}

\begin{biography}{FCS-251369-fig14}
    \textbf{Yucheng Ning} received the B.S. degree in Artificial Intelligence from the University of Chinese Academy of Sciences (UCAS) in 2025, where he is currently pursuing the Ph.D. degree with the Institute of Information Engineering (IIE), Chinese Academy of Sciences. His research interests include natural language processing (NLP) and LLM safety.
\end{biography}

\vspace{\baselineskip}

\begin{biography}{FCS-251369-fig15}
    \textbf{Xixun Lin} received his Ph.D. degree from the Institute of Information Engineering (IIE), Chinese Academy of Sciences in 2022. He is currently an Assistant Professor at IIE. His research focuses on  trustworthy machine learning. He has published over 30 papers in top-tier academic journals and conferences, including TPAMI, TKDE, TNNLS, ICML, KDD, and WWW. 
\end{biography}

\vspace{\baselineskip}

\begin{biography}{FCS-251369-fig16}

    \textbf{Fang Fang} obtained her Ph.D. from the Institute of Computing Technology, CAS in 2019. Her research focuses on intelligent text content processing, including information extraction and knowledge graph in natural language processing. To date, she has published over 20 papers in CCF-A/B conferences like ACL and AAAI. 
\end{biography}

\vspace{\baselineskip}

\begin{biography}{FCS-251369-fig17}
    \textbf{Yanan Cao} obtained a Ph.D. degree from the Institute of Computing Technology, Chinese Academy of Sciences in 2012. She is a professor at the Institute of Information Engineering, Chinese Academy of Sciences. Her research interests include social network analysis and natural language processing. To date, she has published more than 70 papers, including AAAI, NeurIPS, IJCAI, ACL, WWW, and she won the best paper award at PAKDD-20.
\end{biography}

\end{document}